\documentclass[10pt,twocolumn,letterpaper]{article}

\usepackage[pagenumbers]{wacv} 

\usepackage{algorithm}
\usepackage{algpseudocode}
\usepackage{caption}
\usepackage{multirow}
\makeatletter
\renewcommand\paragraph{\@startsection{paragraph}{4}{\z@}%
                                    {1.5ex \@plus0.5ex \@minus.2ex}%
                                    {-1em}%
                                    {\normalfont\normalsize\bfseries}}
\makeatother

\definecolor{wacvblue}{rgb}{0.21,0.49,0.74}
\usepackage[pagebackref,breaklinks,colorlinks,allcolors=wacvblue]{hyperref}

\def\wacvPaperID{204} 
\def\confName{WACV}
\def\confYear{2027}

\title{Spatially Gated Demographic Steering for Debiasing Multi-Person Text-to-Image Generation}
\author{Venkatesh Thirugnana Sambandham\\
Technische Hochschule Ingolstadt\\
Ingolstadt, Germany\\
{\tt\small venkatesh.thirugnanasambandham@thi.de}
\and
Torsten Schön\\
Technische Hochschule Ingolstadt\\
Ingolstadt, Germany\\
{\tt\small torsten.schoen@thi.de}
}

\begin{document}
\maketitle
\begin{abstract}

Text-to-Image models reproduce societal stereotypes, and existing mitigation applies one demographic correction to the whole scene.
In multi-person scene generation this creates a trade-off: global steering improves aggregate balance only by homogenizing the people within an image.
Steering-based methods like Embedding Arithmetic drive the intra-image diversity below the unsteered baseline.
Text- and latent-projection approaches shift the aggregate skew with only marginal gains in intra-image diversity.
We present Spatially Gated Steering (SGS), which steers the demographics of each person independently by extracting per-person masks from the Diffusion Transformer's (DiT) internal attention maps at a critical denoising timestep and applying a distinct demographic vector within each mask for the remaining steps.
SGS needs no fine-tuning, inversion or layout specifications, and uses a lightweight segmentor over frozen DiT attention features to localize subjects at early timesteps.
On skewed occupational prompts, SGS reaches high intra-image racial and gender diversity while keeping aggregate balance competitive and image quality near baseline, a result we validate on two architecturally distinct diffusion transformers, FLUX.1-Dev and SD3.5-Large.
Codebase and model weights are available at \url{https://github.com/cvims/spatially-gated-steering}

\end{abstract}

\section{Introduction}
Text-to-Image (T2I) generation has transformed digital content creation, yet foundational models such as FLUX~\cite{flux2024} and Stable Diffusion~\cite{pmlr-v235-esser24a} systematically reproduce societal stereotypes~\cite{dehdashtianoasis}.
Gender-neutral prompts like \textit{a carpenter} yield predominantly male-presenting images; \textit{a nurse} yields predominantly female-presenting images~\cite{luccioni2023stable}.
Studies show these outputs shift user beliefs about real-world distributions~\cite{guilbeault2024online}, making fairness mitigation essential for responsible deployment.

\begin{figure}[t]
  \centering
  \includegraphics[width=\linewidth]{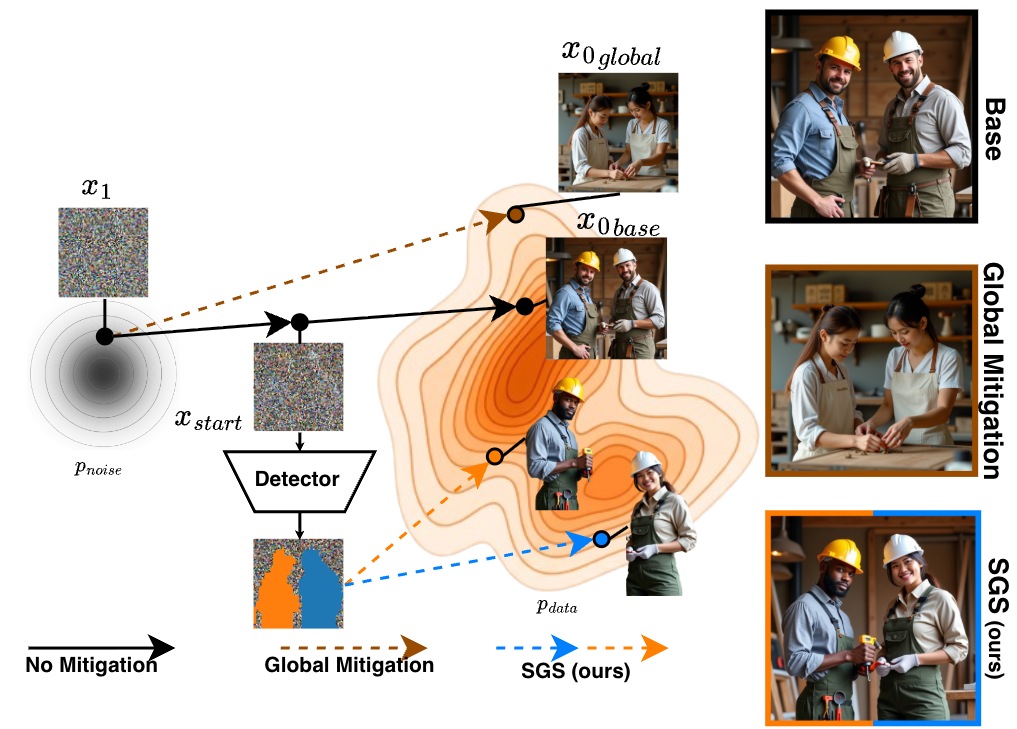}
  \caption{
    \textbf{SGS intuition.}
    The base trajectory (black) reaches $x_0^{\text{base}}$ in a stereotyped region of $p_{\text{data}}$; global mitigation~\cite{EA} (brown dashed) shifts it to one alternative mode.
    SGS instead intercepts at $x_{\text{start}}$, extracts per-person masks, and steers each individual independently (orange/blue dashed) toward distinct demographic attractors within the same image $x_0$.
  }
  \label{fig:intuition}
\end{figure}

Several mitigation strategies have been proposed and they operate at different levels of the diffusion pipeline~\cite{smithoperationalizing}: They modify the text embeddings~\cite{EA,chuang2023debiasing,zhang2023iti,kim2025rethinking} or fine-tune model weights~\cite{gandikota2024unified,shenfinetuning} to shift the aggregate output distribution.
While effective for scenes with a single person, global mitigation methods share a fundamental limitation: They apply identical steering to every person in a scene.
A prompt like "a group of nurses" receives one demographic vector, pushing all the persons in the image in the same direction and failing to diversify individuals independently.

\begin{figure*}[t!]
    \centering
    \includegraphics[width=0.9\linewidth]{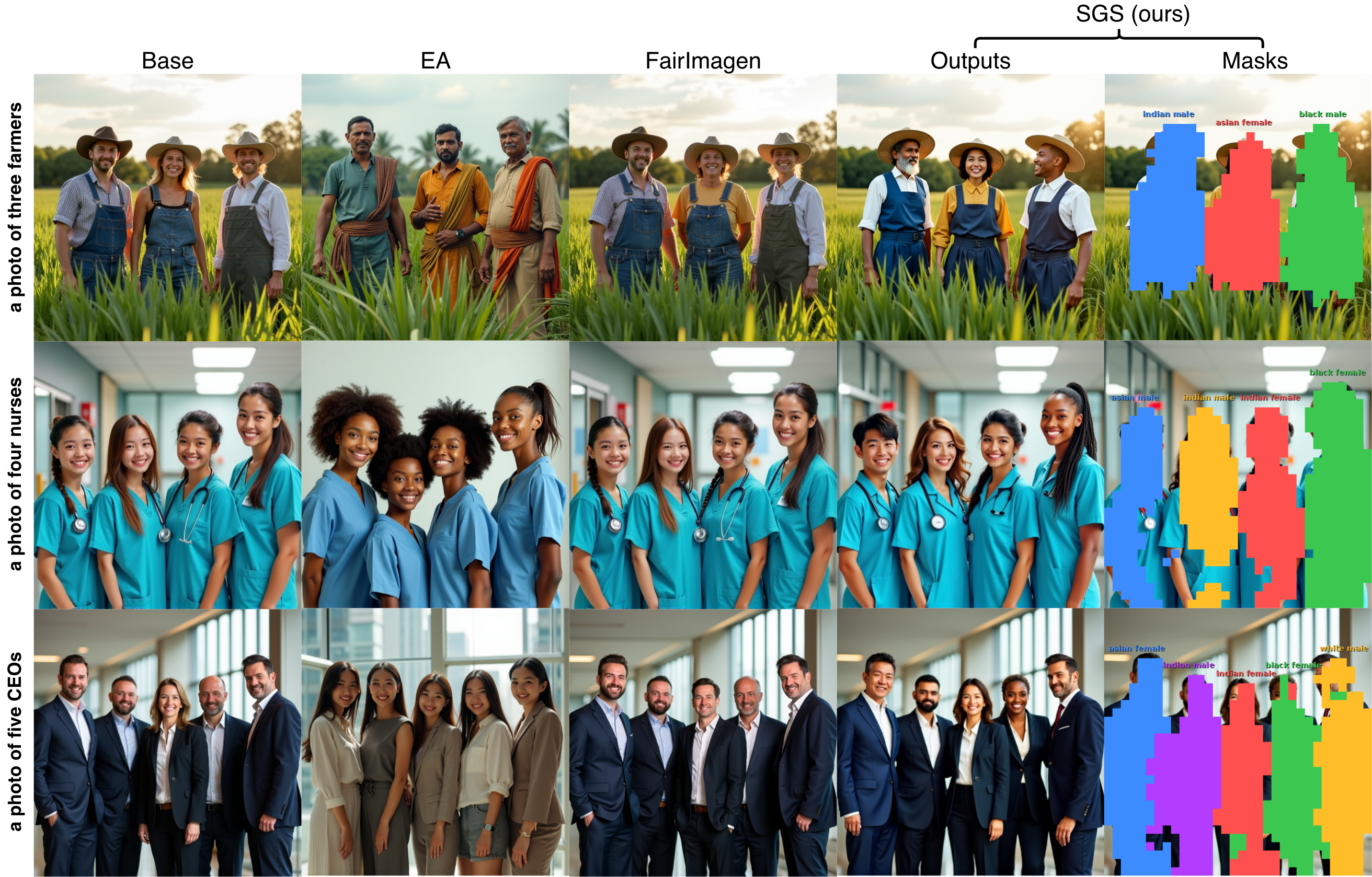}
    \caption{\textbf{Qualitative comparison across three profession types spanning different stereotype directions.} Prompts contain no demographic descriptors. \textbf{Base} is demographically homogeneous; \textbf{EA} shifts everyone uniformly; \textbf{FairImagen} is statistically indistinguishable from baseline on FLUX; \textbf{SGS} steers each instance toward a distinct intersectional target while preserving occupational context and scene coherence.
    Rightmost column: segmentation masks (nearest-neighbor upsampled from $32\times32$) with assigned demographic labels.
    }
    \label{fig:qualitative}
\end{figure*}

This limitation is not merely theoretical: \citet{wan2024pstest} show that multi-person prompts expose a distinct, compounding class of bias that single-subject evaluations cannot capture. Co-occurring subjects amplify stereotypes even when the same model behaves anti-stereotypically on single-person prompts.

Attempts to address this gap remain limited. Existing approaches either rewrite the generation prompt with a single corrective signal applied to the whole scene~\citep{wan2024pstest,davoodi2026kg}, optimize demographic distributions at the aggregate level without per-instance control~\citep{jung2025mpr}, or require an explicit spatial layout as input~\citep{dahary2024yourself}. 
To the best of our knowledge, no existing method localizes individuals from the model's own internal representations and applies independent demographic steering, to achieve intra-image diversity.
Figure~\ref{fig:intuition} illustrates this gap: while global mitigation shifts all persons uniformly toward a single demographic mode, our approach steers each individual independently through their spatial mask, reaching distinct attractors within $p_{\text{data}}$ in a single denoising pass.

We introduce \textbf{Spatially Gated Steering (SGS)}, a framework for per-instance demographic steering in multi-person T2I scene generation.
SGS operates by 1) Identifying and generating the masks of instances of people during the intermediate denoising steps using a lightweight detector, 2) Steering the demographics of the detected individuals, while also maintaining concept coherence.

Our contributions are as follows:

\begin{itemize}

\item \textbf{A trade-off in multi-person fairness.} 
Global mitigation faces an inherent tension: improving aggregate balance requires collapsing intra-image diversity, sometimes below the unsteered baseline.
Since any per-instance diversification trivially raises intra-image entropy, the meaningful result is high entropy at a competitive aggregate balance.
SGS reaches high intra-image diversity with a competitive aggregate balance by steering per-instance without fine-tuning, flow-inversion or layout.
\item \textbf{Instance segmentation from DiT features.} We show that self-attention maps of DiTs at a critical denoising timestep encode spatially separable person instances, and train a lightweight DETR-style~\cite{carion2020end} segmentor on these frozen features that isolates individuals across crowd sizes and scene complexities.
\item \textbf{Benchmark and analysis.} We introduce a benchmark measuring fairness along both intra-image diversity and aggregate deviation, surfacing the homogenization failure mode of global methods and showing SGS is strong on both axes, on both FLUX.1-Dev and SD3.5-Large.

\end{itemize}

\section{Related Works}

\paragraph{Bias and Fairness in Text-to-Image Models.}
T2I models have been extensively shown to amplify demographic stereotypes encoded in training data, producing outputs skewed by gender, race and cultural identity across a wide range of prompt types~\citep{bianchi2023easily, cho2023dall, seshadri2024bias, luccioni2023stable}.
Beyond documenting these skews, a parallel line of work formalizes their measurement: Dehdashtian et al.~\cite{dehdashtianoasis} quantify stereotypical attributes in the generated set and trace their origin within the model, reporting that even high-fidelity systems such as FLUX retain strong stereotypical predispositions.
Mitigation strategies operate at different points in the generation pipeline: prompt engineering injects demographic descriptors at inference~\citep{friedrich2025auditing, bonna2024debiaspi,zhang2023iti}, embedding manipulation steers text representations prior to image generation~\citep{EA, chuang2023debiasing,kim2025rethinking,li2024self,li2025fair,han2026lightfair}, and parameter-efficient fine-tuning edits model weights to shift output distributions~\citep{gandikota2024unified,shenfinetuning}.
Park et al.'s Entanglement-Free Attention~\citep{park2025fair} shares our goal of preserving non-target attributes during bias mitigation, training cross-attention predictors under a mask-constrained regularization loss so background edits are suppressed.
However, like other global methods, it applies one scene-level attribute correction rather than steering each person independently.
Despite this breadth, all existing methods treat the generated image as a single unit, applying one steering signal to the entire scene.
This is sufficient when a prompt produces one person, but breaks down entirely for multi-person scenes where distinct individuals require independent control.

\paragraph{Multi-person Bias and Instance-level Control.}
This breakdown in multi-subject generation is primarily driven by semantic leakage, a phenomenon where the attention mechanisms of diffusion architectures inadvertently blend the visual and demographic features of different subjects during the denoising trajectory. 
\citet{wan2024pstest} demonstrate the consequences of this at scale: their Paired Stereotype Test shows that over 74\% of DALL-E~3 images depicting two individuals with contrastive social identities display gender-occupational biases, even when the same model produces anti-stereotypical outputs for single-person prompts. 
\citet{jung2025mpr} propose a Multi-Group Proportional Representation metric to measure and optimize demographic distributions across intersectional groups, but operate at the aggregate level without per-instance control.
A related direction targets stereotypes that emerge from the co-occurrence of multiple concepts within a single prompt: \citet{zhou2024association} show that pairing a demographic descriptor with an unrelated object can induce stereotypical associations invisible to single-concept audits, mitigated via a trained Sensitive Transformer that constrains the output distribution toward a stereotype-free target. This targets a distinct problem from ours,  stereotyped associations between a subject and its surrounding context, corrected globally over the whole generation, rather than diversifying multiple people already explicitly present in a scene.
Training-free inference methods such as Bounded Attention~\citep{dahary2024yourself} attempt to prevent semantic leakage by confining attention flows within manually predefined spatial bounding boxes, but rely on rigid user-provided layouts rather than organic scene composition. Reinforcement learning approaches such as DisCo~\citep{borse2025disco} directly fine-tune models to penalize intra-image facial similarity, but use face similarity as a proxy for demographic diversity rather than steering specific attribute vectors, and incur significant computational overhead. 
Global steering techniques such as Embedding Arithmetic~\citep{EA} modify text embeddings uniformly, which fundamentally fails to produce heterogeneous intra-image diversity.

\paragraph{Diffusion Features for Spatial Understanding.}
Intermediate diffusion activations encode rich spatial structure without additional training: DIFT~\citep{tang2023emergent} shows mid-denoising U-Net features support zero-shot dense correspondence; ODISE~\citep{xu2023open} uses Stable Diffusion U-Net features for open-vocabulary panoptic segmentation.
This extends to Diffusion Transformers only recently: ConceptAttention~\citep{helblingconceptattention} shows DiT attention-layer projections yield sharper concept saliency than cross-attention alone; TRACE~\citep{jo2025trace} shows diffusion models isolate object boundaries at a specific timestep, enabling spatial masks from internal attention.
SGS extends this line of work to intra-image demographic fairness in multi-person scenes generated by T2I models.

\section{Preliminaries}
\label{sec:prelims}

\subsection{Rectified Flow Denoising}
\label{sec:rf}

Rectified flow~\cite{liuflow} learns a velocity field $v_\theta$ (a neural network with parameters $\theta$) that transports samples from a Gaussian noise distribution to the data distribution along approximately straight trajectories.
FLUX~\cite{flux2024} adopts this formulation, initializing generation from $x_1 \sim \mathcal{N}(0, I)$ and iteratively denoising via:
\begin{equation}
    x_{t - \Delta t} \;=\; x_t \;-\;
    \Delta t \cdot v_\theta(x_t,\, t,\, \mathbf{c}),
    \label{eq:rf}
\end{equation}
where $t \in [0, 1]$ decreases from $1$ (pure noise) to $0$ (clean image) over $T$ steps, $\Delta t = 1/T$, and $\mathbf{c}$ is the encoded text condition~\cite{radford2021learning}.

\subsection{Multimodal Diffusion Transformer (MM-DiT)}
\label{sec:mmdit}

The MM-DiT~\cite{pmlr-v235-esser24a} architecture patchifies the noisy image latent $x_t \in \mathbb{R}^{HW\times d}$, where $HW$ are the latent height and width and $d$ is the token dimension, into image tokens, concatenates them with text conditioning tokens $\mathbf{c}$~\cite{radford2021learning}, and processes them through a stack of $N$ transformer layers, each applying joint image-text self-attention and a feedforward MLP.
FLUX.1-Dev~\cite{flux2024} uses $N{=}19$ double-stream layers; SD3.5-Large~\cite{pmlr-v235-esser24a} uses $N{=}38$ joint transformer blocks.

\subsection{Embedding Arithmetic}
\label{sec:ea}
Embedding Arithmetic~\cite{EA} computes a demographic attribute vector as the mean difference between embeddings generated by a biased prompt ($p_{\mathrm{target}}$) and neutral prompts ($p_{\mathrm{neutral}}$) over a set of profession prompts $\mathcal{P}$:
\begin{equation}
    \mathbf{v}_A = \frac{1}{|\mathcal{P}|}
    \sum_{p \in \mathcal{P}}
    \bigl(
        \mathbf{e}(p_{\mathrm{target}})
        - \mathbf{e}(p_{\mathrm{neutral}})
    \bigr),
    \label{eq:ea}
\end{equation}
where $\mathbf{e} (\cdot) \in \mathbb{R}^D$ denotes the text embedding and $A$ indexes the target demographic attribute.
FLUX conditions denoising on two text encoders: a pooled CLIP-L/14 embedding (768-dim) that modulates the network globally, and a T5-XXL sequence embedding that carries the prompt text via cross-attention. 
Embedding Arithmetic's direction vectors are computed and validated in the CLIP pooled space -- $\mathbf{e}(\cdot)$ and $\mathbf{c}$ in Eq.~(\ref{eq:rf}) denote this pooled CLIP embedding, not the T5 sequence. 
$p_{\mathrm{target}}$ and $p_{\mathrm{neutral}}$ share an identical template except for an inserted demographic descriptor (e.g.\ ``a photo of a \{demographic\} \{profession\}'' vs.\ ``a photo of a \{profession\}''), so $\mathbf{v}_A$ isolates the demographic direction rather than any profession-specific signal.
At inference, the original method shifts the conditioning vector globally as $\tilde{\mathbf{c}} = \mathbf{c} + \alpha \cdot \mathbf{v}_A$, where $\alpha > 0$ controls the steering strength.

\section{Methods}
\label{sec:method}
SGS performs per-instance demographic steering within the standard rectified flow denoising loop, requiring no model fine-tuning, no trajectory inversion, and no external spatial layout specification.
Crucially, instance masks are predicted directly from the MM-DiT's attention features at a critical denoising step, rather than decoding the noisy latents for an external segmentor - which would be unreliable while the latent is still far from $x_0$, and redundant since these features are already computed by the base pass.
Figure~\ref{fig:intuition} illustrates the intuition behind our approach and Algorithm~\ref{alg:sgs} summarizes the complete inference procedure.
The denoising loop is divided into three phases governed by two hyperparameters: $t_{\mathrm{start}}$ and $t_{\mathrm{end}}$.
Before $t_{\mathrm{start}}$, generation proceeds without intervention.
At $t_{\mathrm{start}}$, instance masks are extracted and cached.
Between $t_{\mathrm{start}}$ and $t_{\mathrm{end}}$, a separate steered forward pass is run per detected instance at each denoising step, and the predictions are spatially accumulated.
After $t_{\mathrm{end}}$, standard denoising resumes to completion.

\subsection{Instance Detection}
\label{sec:seg}
At step $t = t_{\mathrm{start}}$, the base forward pass is run with attention extraction enabled, returning the MM-DiT self-attention maps $\{\mathrm{SA}_{t}^\ell\}_{\ell=1}^{N}$.

\paragraph{Self-attention maps.}
The self-attention map for layer $\ell$ is computed independently for each attention head and averaged across heads, capturing pairwise spatial affinities between image tokens~\cite{jo2025trace}; the full expression is given in supplementary Section~\ref{app:segmentor}.

\paragraph{Layer binning.}
In order to save compute and memory usage, rather than operating on all $N$ transformer layers independently, we partition them into $B{=}4$ non-overlapping bins of approximately 5 consecutive layers each for FLUX.1-Dev and 9--11 layers each for SD3.5-Large. 
We then compute the layer-wise mean of the self-attention maps within each bin, yielding four spatially resolved feature maps $\{\widehat{\mathrm{SA}}_t^b\}_{b=1}^{4}$ per timestep.
Through an ablation study over all bin combinations (Section~\ref{sec:seg_results}), we find that concatenating bins $b{=}3$ and $b{=}4$ yields the best instance segmentation performance.
We therefore define the feature representation as:
\begin{equation}
\mathbf{F}_t=
\widehat{\mathrm{SA}}_t^3 \Vert
\widehat{\mathrm{SA}}_t^4
\in \mathbb{R}^{HW \times 2HW},
\label{eq:feat}
\end{equation}
where $\Vert$ denotes channel-wise concatenation.
$\mathbf{F}_t$ is passed directly to the instance segmentor described below.

\begin{algorithm}[t]
\small
\caption{SGS Inference}
\label{alg:sgs}
\begin{algorithmic}[1]
  \Require Noisy image latent $x_1$, prompt embeddings
           $(\mathbf{c})$,
           attribute vectors $\{\mathbf{v}_A\}$,
           strength $\alpha$, $\lambda$,
           window $[t_{\mathrm{start}}, t_{\mathrm{end}})$
  \For{$t = 0, \ldots, T-1$}
    \State $\hat{v}^{\mathrm{base}} \leftarrow
           v_\theta(x_t, t, \mathbf{c})$
           \Comment{base pass}
    \If{$t = t_{\mathrm{start}}$}
      \State $\mathbf{F}_t \leftarrow
             \mathrm{BinAttn}(\{\mathrm{SA}_t^\ell\})$
             \Comment{Eq.~\eqref{eq:feat}}
      \State $\{M_t\}, K \leftarrow
             \mathrm{Segmentor}(\mathbf{F}_t)$
             \Comment{Eq.~\eqref{eq:seg}}
      \State $\{A_t\} \leftarrow
             \mathrm{AssignDemographics}(K)$
    \EndIf
    \If{$t_{\mathrm{start}} \leq t < t_{\mathrm{end}}$
        \textbf{ and } $K > 0$}
      \State $\hat{v}^{\mathrm{final}} \leftarrow
             \hat{v}^{\mathrm{base}}$
      \For{$i = 1, \ldots, K$}
        \State $\tilde{\mathbf{c}}^{(i)} \leftarrow
               \mathbf{c}^{(i)} +
               \alpha \cdot \mathbf{v}_{A_i}$
               \Comment{Eq.~\eqref{eq:ea_shift}}
        \State $\hat{v}^{(i)} \leftarrow
               v_\theta(x_t, t,
               \tilde{\mathbf{c}}^{(i)})$
               \Comment{Eq.~\eqref{eq:fair_pred}}
        \State $\hat{v}^{\mathrm{final}} \mathrel{+}=
               \lambda\,(\hat{v}^{(i)} -
               \hat{v}^{\mathrm{base}}) \odot M_i$
               \Comment{Eq.~\eqref{eq:accum}}
      \EndFor
    \Else
      \State $\hat{v}^{\mathrm{final}} \leftarrow
             \hat{v}^{\mathrm{base}}$
    \EndIf
    \State $x_{t - \Delta t} \leftarrow x_t +
           (t_{\mathrm{prev}} - t_{\mathrm{curr}}) \cdot
           \hat{v}^{\mathrm{final}}$
           \Comment{Eq.~\eqref{eq:update}}
  \EndFor
  \State \Return $x_0$
\end{algorithmic}
\end{algorithm}

\paragraph{Segmentor Architecture.}
$\mathbf{F}_t$ is passed to a DETR-style \cite{carion2020end} instance segmentor.
The model comprises a three-layer CNN backbone that projects $\mathbf{F}_t \in \mathbb{R}^{HW \times 2HW}$ into a spatial feature map, followed by a linear pixel projection that yields a sequence of pixel embeddings $\mathbf{P} \in \mathbb{R}^{HW \times D_e}$ where $D_e$ is the transformer embedding dimension.
A set of $N_q$ learnable instance queries $\mathbf{Q} \in \mathbb{R}^{N_q \times D_e}$ are refined by a $L$-layer transformer decoder attending over $\mathbf{P}$.
The denoising timestep $t$ is incorporated via a sinusoidal embedding concatenated to the input features, conditioning the segmentor on the noise level at which features were extracted.
Each refined query produces three outputs via dedicated linear heads:
\begin{equation}
    \mathrm{conf}_i \in [0,1], \quad
    \mathrm{center}_i \in [0,32]^2, \quad
    M_i \in [0,1]^{32 \times 32},
    \label{eq:seg}
\end{equation}
where $\mathrm{conf}_i$ is the instance confidence, $\mathrm{center}_i$ is the predicted center coordinate, and $M_i$ is the instance mask produced via a dot product between the query mask embedding and the pixel embedding sequence.
Queries with confidence below a threshold $\tau$ are discarded, yielding $K \leq N_q$ active instances.
More details, along with the hyperparameters of the segmentor can be found in the supplementary material (Section \ref{app:segmentor}).

\paragraph{Training Data.}
The segmentor is trained on a purpose-built dataset of $1440$ generated multi-person images spanning $k {\in} \{2, 3, 4, 5\}$ persons across 9 scene types and 3 difficulty levels, with instance pseudo-labels produced by SAM2~\cite{ravi2025sam} automatic mask generation.
Full dataset details and the train/val split are provided in the supplementary material.

\paragraph{Training.} The segmentor is trained with a Hungarian matching loss~\cite{carion2020end}.
For each image, $N_q$ predicted queries are optimally assigned to $K$ ground-truth instances by minimising a cost matrix.
$$\mathcal{C}_{ij} = w_c \frac{|\hat{c}_i - c_j|_2}{32}
   - \log \hat{p}_i
   + w_m\,(1 - \mathrm{Dice}(\hat{M}_i, M_j))$$

where $\hat{c}_i$, $\hat{p}_i$, $\hat{M}_i$ are the predicted center, confidence, and mask for query $i$, $c_j$, $M_j$ are the ground-truth center and mask, and $w_c$, $w_m$ weight the center and mask loss terms (values in supplementary Section~\ref{app:segmentor}).
Unmatched queries are supervised with a no-instance label.
The mask head is trained with Dice loss~\cite{milletari2016v} and binary cross-entropy on the matched pairs.

\paragraph{Mask caching and demographic assignment.}
The $K$ binary masks $\{M_i\}_{i=1}^{K}$ are cached and reused for all subsequent steps in the steering window without re-extraction.
Each instance $i$ is assigned a target demographic attribute $A_i$ by uniform sampling over the intersection of gender and racial attribute categories, which can be overridden at inference time to reflect user-specified fairness targets.

\subsection{Per-Instance Steering}
For each timestep $t$ in the window $t_{\mathrm{start}} \leq t \leq t_{\mathrm{end}}$, the base velocity prediction $\hat{v}_\theta$ is first obtained from the standard forward process:
\begin{equation}
    \hat{v}_{\theta}^{\mathrm{base}} = v_{\theta}(x_t,\, t,\, \mathbf{c})
\end{equation}
where $\mathbf{c}$ is the conditional embeddings from the text encoder.
Rather than the single global shift Embedding Arithmetic applies (Section~\ref{sec:ea}), SGS shifts the conditioning vector independently per detected instance:
\begin{equation}
    \tilde{\mathbf{c}}^{(i)} = \mathbf{c}^{(i)} + \alpha \cdot \mathbf{v}_{A_i},
    \label{eq:ea_shift}
\end{equation}
where $\alpha > 0$ controls the steering strength and $A_i$ is the demographic attribute assigned to the instance $i$.
SGS pre-computes $\{\mathbf{v}_A\}$ offline for a set of gender and racial attributes and applies Eq.~\ref{eq:ea_shift} independently per detected person within the steering window described in Section~\ref{sec:ablation}.

\paragraph{Fair Forward Pass.}
For each instance $i \in \{1, \ldots, K\}$, a separate fair forward pass is run with the conditioning embedding $\tilde{c}$ steered via Eq.~\eqref{eq:ea_shift}:
\begin{equation}
    \hat{v}_\theta^{(i)} = v_\theta\!\left(x_t,\, t,\, \tilde{\mathbf{c}}^{(i)}\right)
    \label{eq:fair_pred}
\end{equation}

\paragraph{Spatially Gated Accumulation}
The per-instance steering vector is the difference between the fair and base predictions:
\begin{equation}
    \delta^{(i)} = \hat{v}_\theta^{(i)} - \hat{v}_\theta^{\mathrm{base}}
\end{equation}
The steered velocity is accumulated across all $K$ instances, with each steering vector spatially gated by the corresponding cached mask $M_i$ of the instance:
\begin{equation}
    \hat{v}_\theta^{\mathrm{final}} = \hat{v}_\theta^{\mathrm{base}} + \lambda \sum_{i = 1}^{K}\delta^{(i)} \odot M_i
    \label{eq:accum}
\end{equation}
where $\lambda > 0$ is the diversity strength and $\odot$ denotes the element-wise multiplication with $M_i$ broadcast over the feature dimension.
The latent is then updated via Eq.~\eqref{eq:rf}.
\begin{equation}
    x_{t - \Delta t} = x_t +
    (t_{\mathrm{prev}} - t_{\mathrm{curr}}) \cdot
    \hat{v}_\theta^{\mathrm{final}}.
    \label{eq:update}
\end{equation}
Outside the steering window, the latent is updated using the base prediction only and no attention extraction, fair forward pass, or mask application occurs.

\section{Experiments}
\label{sec:experiments}
\subsection{Implementation Details}
\label{sec:impl}
\paragraph{Model and Hardware Configurations.}
Our primary experiments use FLUX.1-Dev\footnote{https://huggingface.co/black-forest-labs/FLUX.1-dev} in bfloat16 precision \cite{kalamkar2019study} on a single NVIDIA A6000 GPU (48GB VRAM), with 21 denoising steps and a guidance scale of 4.5.
To evaluate cross-architecture generalization, we additionally port SGS and all four baselines to SD3.5-Large\footnote{https://huggingface.co/stabilityai/stable-diffusion-3.5-large} in bfloat16 precision, run on a single NVIDIA A100 GPU (39GB MIG slice), at $1024{\times}1024$ resolution (its native training resolution) with 28 denoising steps and a guidance scale of 3.5.
SGS's instance-detection and per-instance steering mechanism (Section~\ref{sec:seg}, Section~\ref{sec:ea}) makes no FLUX-specific assumption beyond the existence of a joint image-text self-attention map inside the denoising transformer, which SD3.5's MM-DiT also provides.
SD3.5-Large's EA direction vectors live in the joint pooled embedding of its two CLIP encoders (CLIP-L and CLIP-G, concatenated to 2048-dim), rather than FLUX's single 768-dim CLIP-L pooled space; T5 remains excluded from vector construction in both cases.

\begin{figure}[t]
    \centering
    \includegraphics[width=\linewidth]{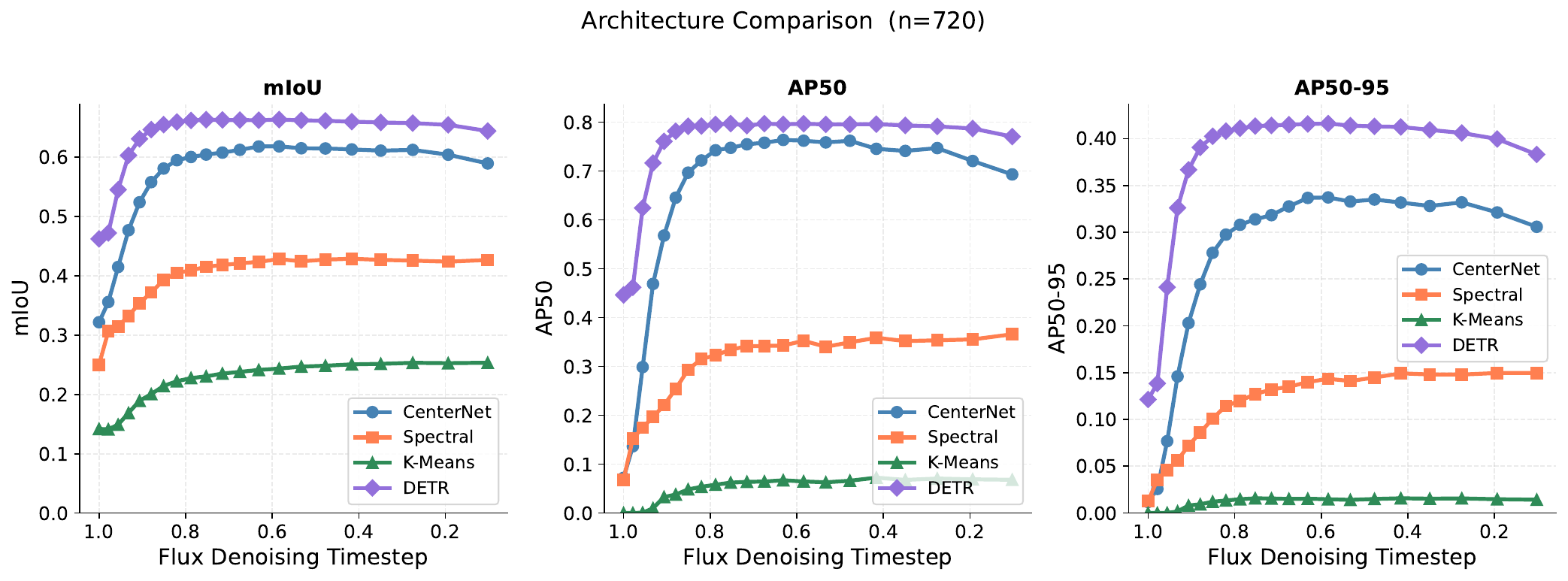}
    \caption{
        \textbf{Instance segmentation analysis} ($n{=}720$):
        segmentation performance across the FLUX denoising
        trajectory for four architectures. All methods peak
        around $t{\approx}0.7$--$0.8$; DETR consistently
        outperforms all baselines. Layer bin ablation motivating
        our choice of Bins 3+4 as $\mathbf{F}_t$ (Eq.~\eqref{eq:feat})
        is in supplementary Section~\ref{app:segmentor}.
    }
    \label{fig:arch_sweep}
\end{figure}

\subsection{Evaluation Metrics}
\paragraph{Intersectional Entropy ($H_{ix}$)}
Let $\mathcal{G}{=}\mathcal{R} \times \mathcal{S}$ denote the set of intersectional demographic groups over racial categories $\mathcal{R}{=}\{\text{white}, \text{black}, \text{asian}, \text{indian}\}$ and genders $\mathcal{S} = \{\text{male}, \text{female}\}$, giving $|\mathcal{G}| = 8$ groups\footnote{We acknowledge that Broader gender and racial spectra exist; we restrict to these for classifier reliability and existing dataset label coverage.}.
For a single generated image with $n$ detected faces, let $p_g$ be the ratio of faces classified as group $g$.
We compute per-image intersectional entropy as:
\begin{equation}
    H_{ix} = -\frac{1}{\log \min(n,\,|\mathcal{G}|)}
    \sum_{g \in \mathcal{G}} p_g \log p_g \;\in [0, 1],
    \label{eq:hix}
\end{equation}
and report its mean over all generated images.
Normalizing the entropy with $\log \min(n, |\mathcal{G}|)$ ensures that a $k{=}2$ image in which both persons belong to a distinct intersectional groups achieves $H_{ix}{=}1.0$, independent of $k$.
This captures \emph{intra-image} demographic diversity which is the core objective of Spatially Gated demographic Steering, rather than aggregated statistics.

\paragraph{Deviation Ratio ($DR$)}
Following~\cite{respodiff,li2024self}, for an attribute with category set $\mathcal{C}$ let $N_c$ be the total number of detected faces classified as category $c$ and $N=\sum_c N_c$:
\begin{equation}
    DR = \max_{c \in \mathcal{C}} \;
    \frac{\bigl|\,N_c/N - 1/|\mathcal{C}|\,\bigr|}
         {1 - 1/|\mathcal{C}|},
    \label{eq:delta}
\end{equation}
where $DR{=}0$ is perfectly balanced and $DR{=}1$ is maximally skewed.
We report $DR_{\text{race}}$ ($|\mathcal{C}|{=}4$: white, black, asian, indian) and $DR_{\text{gender}}$ ($|\mathcal{C}|{=}2$: male, female).
Unlike the original single-person formulation, we aggregate face-level counts across all $k$ persons per image, which we find more appropriate for multi-person generation.

\paragraph{Image Quality}
We report CLIP score~\cite{radford2021learning} for prompt fidelity and aesthetic score delta ($Aes\Delta$) relative to the unsteered baseline~\cite{schuhmann2022laion}.
Face crops from the generated images are extracted using MTCNN~\cite{mtcnn}.
Similar to prior works~\citep{EA,wan2024pstest}, Demographic labels are produced by FairFace~\cite{karkkainen2019fairface} detector, which we validate directly against known ground truth rather than assuming its accuracy (supplementary Section~\ref{app:classifier_validation}).

\subsection{Instance Segmentation}
\label{sec:seg_results}
\paragraph{Evaluation Protocol.}
We evaluate instance segmentation on a held-out test set of our dataset as described in Section~\ref{sec:seg}.
The test set has $n{=}720$ images across all $k\in\{2,3,4,5\}$ person counts and three difficulty levels.
We report mean intersection-over-union (mIoU), average precision at IoU threshold (AP50) and average precision averaged over thresholds 50\%--95\% (AP50-95).
Features are extracted at each denoising timestep $t \in [0, 1]$ and we report performance across the full temporal sweep.

\paragraph{Architectural Comparison}
We compare four segmentation approaches of increasing supervisory complexity: K-Means (naive unsupervised baseline), Spectral clustering (following MaskDiffusion~\cite{kawano2024maskdiffusion}), CenterNet~\cite{zhou2019objects} (supervised, lightweight), and our DETR-style model (Section~\ref{sec:seg}).
TRACE~\cite{jo2025trace}, which also extracts instance boundaries from diffusion self-attention maps, is a natural comparison, but its authors had not released training or inference code at submission time, precluding a faithful reproduction.

Figure~\ref{fig:arch_sweep} reports performance across FLUX denoising timesteps.
All methods improve as denoising progresses toward $t{\approx}0.7$--$0.8$ (step 4 of 21), then plateau or degrade as features shift toward fine-grained texture, confirming MM-DiT features become instance-discriminative at intermediate timesteps~\cite{jo2025trace,tang2023emergent}; together with the steering-window ablation (Section~\ref{sec:ablation}), this motivates $t_{\mathrm{start}}{=}4$.
DETR outperforms all baselines at every timestep, peaking at mIoU \textbf{0.664}, AP50 \textbf{0.797}, AP50--95 \textbf{0.416} (vs.\ CenterNet's 0.618/0.762/0.337), reflecting the gap between supervised and unsupervised instance separation from DiT features.
A DETR trained analogously on SD3.5-Large reaches a comparable peak (Section~\ref{app:segmentor_sweep_sd3}), so this timestep-localized discriminability is not FLUX-specific.
Per-difficulty and qualitative detail: supplementary Sections~\ref{app:segmentor_per_difficulty}, \ref{app:segmentor_qualitative_results}; Section~\ref{app:person_count_oracle} shows the segmentor is not the bottleneck behind SGS's steering accuracy at $K{\le}3$, but shows weaker performance at $K{=}4$.

\paragraph{Layer Bin Ablation.}
Performance improves monotonically from Bin 1 to Bin 4, reflecting the progression of spatial structure across MM-DiT layer depth - earlier layers encode coarse layout while later layers capture fine-grained instance boundaries.
Concatenating Bins 3 and 4 achieves the best performance across all metrics (mIoU: 0.631, AP50: 0.747, AP50-95: 0.368, averaged across all timesteps), outperforming other bin configurations (full per-bin table in supplementary Section~\ref{app:segmentor}).
This result directly motivates our choice of $\mathbf{F}_t = \widehat{\mathrm{SA}}_t^3 \Vert \widehat{\mathrm{SA}}_t^4$ in Eq.~\eqref{eq:feat}.

\subsection{Steering Parameters Ablation}
\label{sec:ablation}
We evaluate on a set of 8 occupations: 4 male-skewed (\textit{lawyer, physician, developer, politician}) and 4 female-skewed (\textit{social worker,  hairdresser, cashier, receptionist}), generating 10 images per occupation for $k{\in}\{2,3\}$ persons with randomly assigned target demographics per detected instance.
We report five metrics: race and gender steering accuracy ($\mathrm{Acc}_{\mathrm{race}}$, $\mathrm{Acc}_{\mathrm{gender}}$) against the assigned target; aesthetic score delta ($Aes\Delta$)~\cite{schuhmann2022laion} relative to the unsteered baseline; 
background preservation ($\mathrm{LPIPS}_{\mathrm{bg}}$~\cite{zhang2018unreasonable}) outside the instance masks; and prompt--image CLIP similarity delta ($\mathrm{CLIP}\Delta$), where values closer to zero indicate less semantic drift.

\paragraph{Steering Hyperparameters.}
We ablate EA strength $\alpha \in \{1.5, 2.0, 2.5, 3.0, 3.5\}$ and spatial diversity strength $\lambda \in \{3.0, 4.0, 5.0, 6.0, 7.0\}$, reporting steering accuracy, aesthetic delta, background preservation, and CLIP similarity delta. 
We select $\alpha{=}2.0$ and $\lambda{=}4.0$ as our steering parameters; full sensitivity analysis is provided in Table~\ref{tab:steering_ablation_supp} of the supplementary material.

\begin{table*}[t]
\centering
\small
\caption{\textbf{Quantitative results on the extended WinoBias occupation benchmark} (30 occupations $\times$ 5 templates $\times$ $k\in\{2,3,4\}$ $\times$ 10 seeds, $n{=}4{,}500$/method/architecture), on FLUX.1-Dev and SD3.5-Large. $H_{ix}$: intra-image intersectional entropy. DR: deviation ratio from uniform (lower is better). \textbf{Bold} and \underline{Underlined} values represent the best and second-best methods per metric, within each architecture.}
\label{tab:benchmark}
\setlength{\tabcolsep}{3pt}
\resizebox{\textwidth}{!}{%
\begin{tabular}{lccccc ccccc}
\toprule
& \multicolumn{5}{c}{FLUX.1-Dev} & \multicolumn{5}{c}{SD3.5-Large} \\
\cmidrule(lr){2-6} \cmidrule(lr){7-11}
Method &
$H_{ix}$ $\uparrow$ & CLIP $\uparrow$ & $Aes\Delta$ $\uparrow$ & DR$_\text{race}$ $\downarrow$ & DR$_\text{gender}$ $\downarrow$ &
$H_{ix}$ $\uparrow$ & CLIP $\uparrow$ & $Aes\Delta$ $\uparrow$ & DR$_\text{race}$ $\downarrow$ & DR$_\text{gender}$ $\downarrow$ \\
\midrule
Baseline & 0.524$\pm$0.016 & \textbf{0.321$\pm$0.001} & -- & 0.748$\pm$0.015 & 0.608$\pm$0.022 & 0.549$\pm$0.015 & \underline{0.325$\pm$0.001} & -- & 0.615$\pm$0.018 & 0.367$\pm$0.026 \\
EA~\citep{EA} & 0.285$\pm$0.013 & 0.236$\pm$0.002 & \textbf{+0.336$\pm$0.011} & \textbf{0.079$\pm$0.016} & \textbf{0.434$\pm$0.027} & 0.319$\pm$0.012 & 0.323$\pm$0.001 & \textbf{+0.163$\pm$0.010} & \textbf{0.069$\pm$0.016} & 0.303$\pm$0.029 \\
SDID~\citep{li2024self} & 0.515$\pm$0.016 & 0.314$\pm$0.001 & \underline{+0.118$\pm$0.008} & 0.797$\pm$0.013 & 0.593$\pm$0.023 & 0.573$\pm$0.015 & 0.322$\pm$0.001 & +0.006$\pm$0.015 & 0.628$\pm$0.017 & 0.456$\pm$0.026 \\
RespoDiff~\citep{respodiff} & \underline{0.540$\pm$0.017} & 0.315$\pm$0.001 & $-$0.055$\pm$0.006 & 0.737$\pm$0.017 & 0.624$\pm$0.022 & \underline{0.575$\pm$0.015} & \textbf{0.326$\pm$0.001} & $-$0.042$\pm$0.009 & 0.584$\pm$0.018 & 0.458$\pm$0.025 \\
FairImagen~\citep{fu2026fairimagen} & 0.534$\pm$0.016 & \underline{0.320$\pm$0.001} & +0.004$\pm$0.005 & 0.744$\pm$0.015 & 0.574$\pm$0.025 & 0.539$\pm$0.015 & 0.319$\pm$0.001 & \underline{+0.057$\pm$0.011} & 0.385$\pm$0.019 & \underline{0.293$\pm$0.028} \\
\midrule
SGS (ours) & \textbf{0.794$\pm$0.014} & 0.318$\pm$0.001 & $-$0.000$\pm$0.007 & \underline{0.299$\pm$0.007} & \underline{0.504$\pm$0.025} & \textbf{0.832$\pm$0.011} & 0.324$\pm$0.001 & $-$0.037$\pm$0.010 & \underline{0.306$\pm$0.006} & \textbf{0.233$\pm$0.028} \\
\bottomrule
\end{tabular}%
}
\end{table*}

\paragraph{Steering Window.}
We additionally sweep detection/steering start step $t_{\mathrm{start}} \in \{2,4,6\}$ and end step $t_{\mathrm{end}}$, with $\alpha{=}2.0$ and $\lambda{=}4.0$ fixed (full grid and discussion in supplementary Section~\ref{app:steering_hyperparameter}, Table~\ref{tab:window_ablation_supp}).
Early detection ($t_{\mathrm{start}}{=}2$) yields the highest steering accuracy but degrades background preservation and aesthetic quality, since MM-DiT features this early are insufficiently structured for reliable instance segmentation.
Late detection ($t_{\mathrm{start}}{=}6$) preserves background coherence but leaves too short a window to shift demographics effectively.
We select $t_{\mathrm{start}}{=}4$, $t_{\mathrm{end}}{=}10$ as our operating point for FLUX, balancing steering accuracy against background preservation and semantic drift.
Similar Hyperparameter testing was done for SD3.5-Large and the results can be found in Section~\ref{app:steering_hyperparameter_sd3}.

\subsection{Benchmark Comparison}
\label{sec:benchmark}
We evaluate on 30 held-out WinoBias occupations (disjoint from the 10 used to fit EA's direction vectors) $\times$ 5 prompt templates $\times$ $k\in\{2,3,4\}$ $\times$ 10 seeds, on both FLUX.1-Dev and SD3.5-Large (Section~\ref{sec:impl}; full list in supplementary Section~\ref{app:extended_benchmark_protocol}), yielding 4{,}500 images/method/architecture.
We compare against Embedding Arithmetic~\cite{EA}, SDID~\cite{li2024self}, RespoDiff~\cite{respodiff}, and FairImagen~\cite{fu2026fairimagen}; DisCo~\cite{borse2025disco} targets a related goal but requires fine-tuning and has no public implementation, precluding a faithful comparison. Each baseline was originally designed for UNet-based diffusion models; how we faithfully reimplemented them for MM-DiT is detailed in supplementary Section~\ref{app:baseline_adaptation}.
$H_{ix}$, CLIP, and $Aes\Delta$ are mean $\pm 95\%$ CI (normal approximation); DR uses a percentile bootstrap (1000 resamples), since it is a population-level statistic rather than a per-image mean.

Table~\ref{tab:benchmark} reports intra-image diversity ($H_{ix}$) against aggregate balance ($DR$) on both architectures under one protocol.
Since $H_{ix}$ rewards any per-instance diversification, the two must be read jointly: global methods steer every person identically and so cannot raise intra-image diversity.
Embedding Arithmetic reaches the lowest $DR_{race}$ on both models, but its $H_{ix}$ falls well below the unsteered baseline on both -- a single demographic vector pushes everyone toward the same group, improving aggregate counts while erasing intra-image variation.
SDID, RespoDiff, and FairImagen leave $DR$ close to baseline on both architectures, confirming that global latent and text-embedding projections do not transfer to either model's joint-attention formulation.
SGS is the only method to break this trade-off on both: highest $H_{ix}$ by a wide margin, $DR_{race}$ second only to EA, and prompt fidelity near baseline throughout.
This pattern replicates on a second, architecturally distinct diffusion transformer, not just FLUX.

\subsection{Human Evaluation}
\label{sec:human_eval}
The metrics above rely on an automatic race/gender classifier (Section~\ref{sec:impl}). To check that SGS's gains are also visible to real people, not just to one automatic tool, we ran a small anonymous online study. Raters were shown two images side by side, one from SGS and one from a baseline, and asked which has more diverse people and which better matches the prompt, plus an overall preference. We collected 65 completed sessions (657 screens, 1971 judgments in total), comparing SGS against all five baselines.

Table~\ref{tab:human_eval} shows the share of judgments that favored SGS. Raters preferred SGS on diversity by a wide margin against every baseline (88--90\%, all differences statistically significant). On matching the prompt, results are mostly a tie, except SGS is clearly preferred over EA (89\%). The overall preference question shows the same pattern: SGS is clearly favored over EA (76\%) and roughly tied with the rest. Agreement between raters (Krippendorff's alpha, where 1.0 means perfect agreement and 0 means no better than chance) was 0.73 for diversity -- strong agreement -- and lower, 0.43--0.44, for the more subjective faithfulness/preference questions, which is expected for personal-taste judgments. Full study design, recruiting, and per-question detail are in the supplementary material (Section~\ref{sec:supp-human-eval}).

\begin{table}[t]
\centering
\small
\caption{\textbf{Human evaluation}: share of judgments preferring SGS over each baseline ($n{=}65$ raters). $^{*}$: statistically significant preference ($p{<}0.05$).}
\label{tab:human_eval}
\begin{tabular}{lccc}
\toprule
vs.\ & Diversity & Faithful. & Preference \\
\midrule
Baseline    & 88.5\%$^{*}$ & 46.2\%       & 47.4\%       \\
EA          & 90.5\%$^{*}$ & 89.4\%$^{*}$ & 75.6\%$^{*}$ \\
FairImagen  & 88.8\%$^{*}$ & 45.8\%       & 46.8\%       \\
RespoDiff   & 89.0\%$^{*}$ & 51.9\%       & 51.3\%       \\
SDID        & 90.5\%$^{*}$ & 52.3\%       & 55.1\%       \\
\bottomrule
\end{tabular}
\end{table}

\subsection{Qualitative Analysis}
Figure~\ref{fig:qualitative}'s three occupations span orthogonal stereotype directions (nurses female-skewed; CEOs, farmers male-skewed), yet base outputs are homogeneous in all three.
EA's unifordm shift is clearest for nurses, where all four figures converge on one demographic, eliminating the intra-image variation that drives its $H_{ix}$ below baseline.
SGS instead maintains scene coherence across all three directions, confining steering to person regions rather than leaking into occupational cues like attire or background.
Additional results are in the supplementary material (Section~\ref{app:supp_qualitative_sec}).

\begin{table}[h]
\centering
\caption{\textbf{Inference time per-image }(A6000 GPU, 512$\times$512, 21 denoising steps). FPs: total forward passes through the DiT per image. Overhead relative to unsteered baseline.}
\label{tab:inference_time}
\begin{tabular}{lccc}
\hline
Method      & \multicolumn{1}{l}{FPs} & \multicolumn{1}{l}{Time (s)} & \multicolumn{1}{l}{Overhead} \\ \hline
Baseline    & 21                      & 7.8 $\pm$ 0.1                    & 1.0                          \\
EA          & 21                      & 7.9 $\pm$ 0.0                    & 1.0                          \\
SGS (K=2) & 33                      & 12.3 $\pm$ 0.0                   & 1.6                          \\
SGS (K=3) & 39                      & 14.6 $\pm$ 0.0                   & 1.9                          \\
SGS (K=4) & 45                      & 16.8 $\pm$ 0.0                   & 2.2                          \\ \hline
\end{tabular}
\end{table}

\subsection{Compute Cost}
Table~\ref{tab:inference_time} reports per-image denoising time measured on a single A6000 GPU (48GB) across 10 runs after 3 warm-up passes.
EA introduces no additional forward passes and is within measurement noise of the baseline.
SGS incurs a deterministic overhead of exactly $K \times (t_{\text{end}}-t_{\text{start}})=6K$ additional forward passes.
For $K=3$ the generation takes 14.6s versus 7.8s for the baseline.
This cost is proportional to the scene complexity and can be reduced by narrowing the steering window at the expense of steering accuracy (Table \ref{tab:window_ablation_supp}, supplementary Section~\ref{app:steering_hyperparameter}).
\section{Limitations}
\label{sec:limitation}
Three limitations bound the current system. 
First, segmentor accuracy degrades for $k{\geq}4$ persons in heavily occluded scenes, this also impacts the steering accuracy.
Second, the per-instance forward pass introduces a compute overhead proportional to $K$. 
Batching would reduce wall-clock time at the cost of VRAM, a trade-off we could not explore within our hardware budget.
Third, SGS applies steering regardless of prompt intent; a query such as ``four male doctors'' will still redistribute demographics. A lightweight prompt parser that zeroes the steering signal for explicitly constrained slots could address this.

\section{Conclusion}
\label{sec:conclusion}

We presented SGS, a per-instance demographic steering framework for multi-person T2I generation requiring no fine-tuning, inversion, or layout input.
Global steering improves aggregate balance only by driving intra-image diversity below the unsteered baseline; SGS is the only evaluated method to break this trade-off, raising intersectional entropy to 0.794 (from 0.524) on FLUX.1-Dev and to 0.832 (from 0.549) on SD3.5-Large, while recovering near-best aggregate balance and preserving semantic coherence on both architectures.
Per-instance spatial steering thus emerges as a necessary step toward fair multi-person text-to-image generation, generalizing across architecturally distinct diffusion transformers rather than being a property of any single model.

{
    \small
    \bibliographystyle{ieeenat_fullname}
    \bibliography{main}
}

\clearpage
\appendix
\setcounter{section}{0}
\renewcommand{\thesection}{\Alph{section}}
\renewcommand{\thetable}{S\arabic{table}}
\setcounter{table}{0}

\section*{Supplementary Material}

\section{Segmentor - Training Dataset Details}
\label{app:segmentor_training}
The segmentor is trained on a purpose-built dataset of 1440 generated multi-person images spanning $k\in\{2,3,4,5\}$ persons across 9 scene types and 3 difficulty levels.
The 9 scene types are: \textit{market, bus stop, city landmark, outdoor concert, train station, backyard barbecue, city park, street food fair,} and \textit{mountain trail}.
Difficulty encodes camera framing and occlusion: \textit{easy} uses a wide shot with subjects standing apart facing the camera;
\textit{medium} uses a medium shot with subjects side by side;
\textit{hard} uses tight close-up shots with partial occlusion among subjects.
All images were generated using FLUX.1-Dev at $512 \times 512$ resolution with guidance scale 4.5 and 21 denoising steps.

\paragraph{Train/val/test split}
The 1440 training images are split into 80\% training and 20\% validation images, stratified by $k$ and difficulty level.
The test set ($n{=}720$) is generated from a disjoint set of 9 out-of-distribution scene types with different framing protocol to evaluate generalisation beyond the training distribution.

\paragraph{Pseudo-label generation}
Instance segmentation masks are produced by SAM2~\cite{ravi2025sam} automatic mask generation applied to each generated image.
Predicted masks are filtered by minimum area and matched to ground-truth person count $k$ via Hungarian matching on mask centroid distance.
Duplicate predictions with pairwise IoU $>0.5$ are suppressed by retaining the higher confidence mask.

\section{Segmentor - Architecture}
\label{app:segmentor}
We describe each component of the instance segmentor with exact hyper-parameter values as used in all reported experiments.
\paragraph{Input Features.}
The segmentor receives the binned self-attention feature map $\mathbf{F}_t$, formed by concatenating layer bins $b{=}3$ and $b{=}4$ of the MM-DiT self-attention at the detection timestep $t_{\mathrm{start}}$.
Features are min-max normalized per channel before being passed to the backbone.
This choice of bins is itself the result of an ablation (Table~\ref{tab:bin_ablation}, moved here from the main paper's Section~5.3 for space): performance improves monotonically from Bin 1 to Bin 4 as layer depth increases, and concatenating Bins 3+4 outperforms every single-bin or other-pair configuration tested.

\begin{table}[h]
\centering
\small
\caption{\textbf{Layer bin ablation} ($n{=}720$). Mean mIoU, AP50, and AP50-95 averaged across all denoising timesteps using DETR. Best in \textbf{bold}.}
\label{tab:bin_ablation}
\setlength{\tabcolsep}{5pt}
\begin{tabular}{lccc}
    \toprule
    \textbf{Bin} & \textbf{mIoU} & \textbf{AP50} & \textbf{AP50-95} \\
    \midrule
    Bin 1 & 0.539 & 0.581 & 0.225 \\
    Bin 2 & 0.583 & 0.661 & 0.287 \\
    Bin 3 & 0.606 & 0.710 & 0.333 \\
    Bin 4 & 0.626 & 0.736 & 0.360 \\
    \midrule
    3+4   & \textbf{0.631} & \textbf{0.747} & \textbf{0.368} \\
    \bottomrule
\end{tabular}
\end{table}

\paragraph{Timestep Conditioning}
The continuous FLUX timestep $t \in [0,1]$ is encoded via a sinusoidal embedding followed by a learned linear projection with SiLU activation. 
The embedding is spatially broadcast and concatenated channel-wise to $\mathbf{F}_t$ before the backbone, conditioning the segmentor on the noise level at which features were extracted.

\paragraph{CNN Backbone}
Three ConvBnReLU blocks (Conv2d, BatchNorm2d, ReLU, same-padding) progressively compress the high-dimensional attention channels into a compact spatial feature map while preserving the $32{\times}32$ resolution throughout. 
A final $1{\times}1$ convolution projects to the transformer embedding dimension $D_e{=}128$, yielding pixel embeddings $\mathbf{P} \in \mathbb{R}^{1{,}024 \times 128}$ after spatial flattening.

\paragraph{Transformer decoder.}
$N_q{=}8$ learnable instance queries are refined by $L{=}2$ decoder layers. 
Each layer applies self-attention among the queries, cross-attention from queries to $\mathbf{P}$, and a two-layer FFN with GELU activation, all with residual connections and LayerNorm ($H{=}4$ attention heads throughout).

\paragraph{Output heads.}
Each refined query produces three outputs via dedicated linear heads: a scalar confidence $\mathrm{conf}_i \in [0,1]$ a center coordinate $\mathrm{center}_i \in [0,32]^{2}$, and an instance mask $M_i \in [0,1]^{32\times32}$ computed as a dot-product between a per-query mask embedding and $\mathbf{P}$, reshaped to $32\times32$.
$\mathrm{center}_i$ is used only during training to stabilize the Hungarian matching.

\paragraph{Inference}
Queries with $\mathrm{conf}_i < \tau{=}0.5$ are discarded and the retained masks are binarized at $> 0.5$, yielding $K \leq 8$ binary instance masks.
No NMS is applied - $K$ is determined by the confidence threshold.

\paragraph{Training.}
The model is trained for 50 epochs with Adam ($\mathrm{lr}{=}10^{-4}$, batch size 8, 5 linear warm-up epochs).
The 1{,}440 training images are split 80/20 into 1{,}152 training and 288 validation samples, stratified by $k$ and difficulty.
Loss weights: $w_c{=}2.0$, $w_m{=}2.0$.
The complete segmentor has 1.2M parameters, lightweight relative to the FLUX DiT backbone it operates on.

\begin{figure*}[t]
    \centering
    \includegraphics[width=\linewidth]{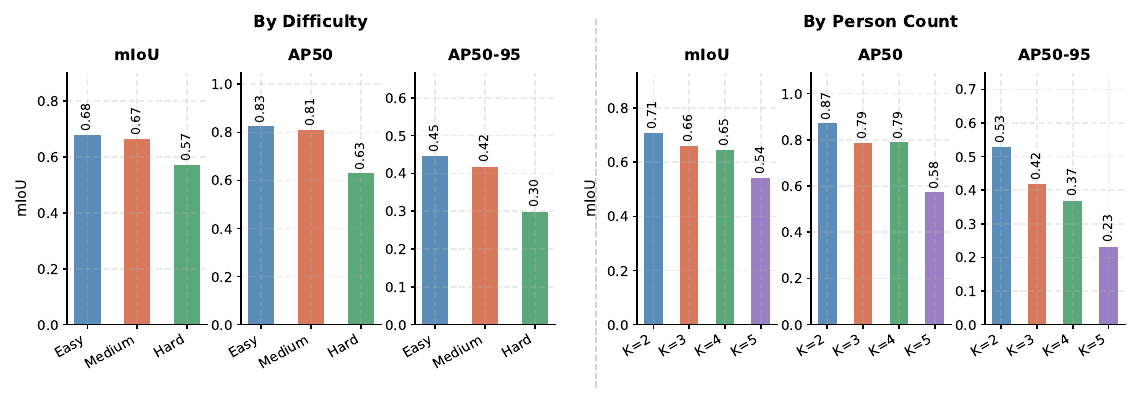}
    \caption{
        Segmentation performance of the SGS instance segmentor (DETR variant) broken down by scene difficulty (\textit{left}) and person count $K$ (\textit{right}).
        \textit{Left:} mIoU, AP50, and AP50-95 averaged over all $K$ for easy, medium, and hard scenes. Performance degrades with difficulty primarily due to increased occlusion and smaller per-person regions.
        \textit{Right:} The same metrics averaged over difficulty levels per $K$. 
        All three metrics decline as $K$ increases, reflecting the greater spatial competition among instances at higher person counts.
        Results are reported at denoising step $t=4$.
    }
    \label{app:seg_difficulty_k}
\end{figure*}

\section{Segmentor Results - Per Difficulty}
\label{app:segmentor_per_difficulty}
Figure~\ref{app:seg_difficulty_k} breaks down DETR segmentor performance across scene difficulty (\textit{left}) and person count $K$ (\textit{right}).
AP50 remains consistently high across all cases, while AP50-95 drops substantially, indicating that instances are reliably detected but mask boundaries lack the precision required at stricter IoU thresholds.
This gap widens at higher $K$, where spatial competition between instances further degrades boundary delineation.

\begin{figure*}[t]
    \centering
    \includegraphics[width=\linewidth]{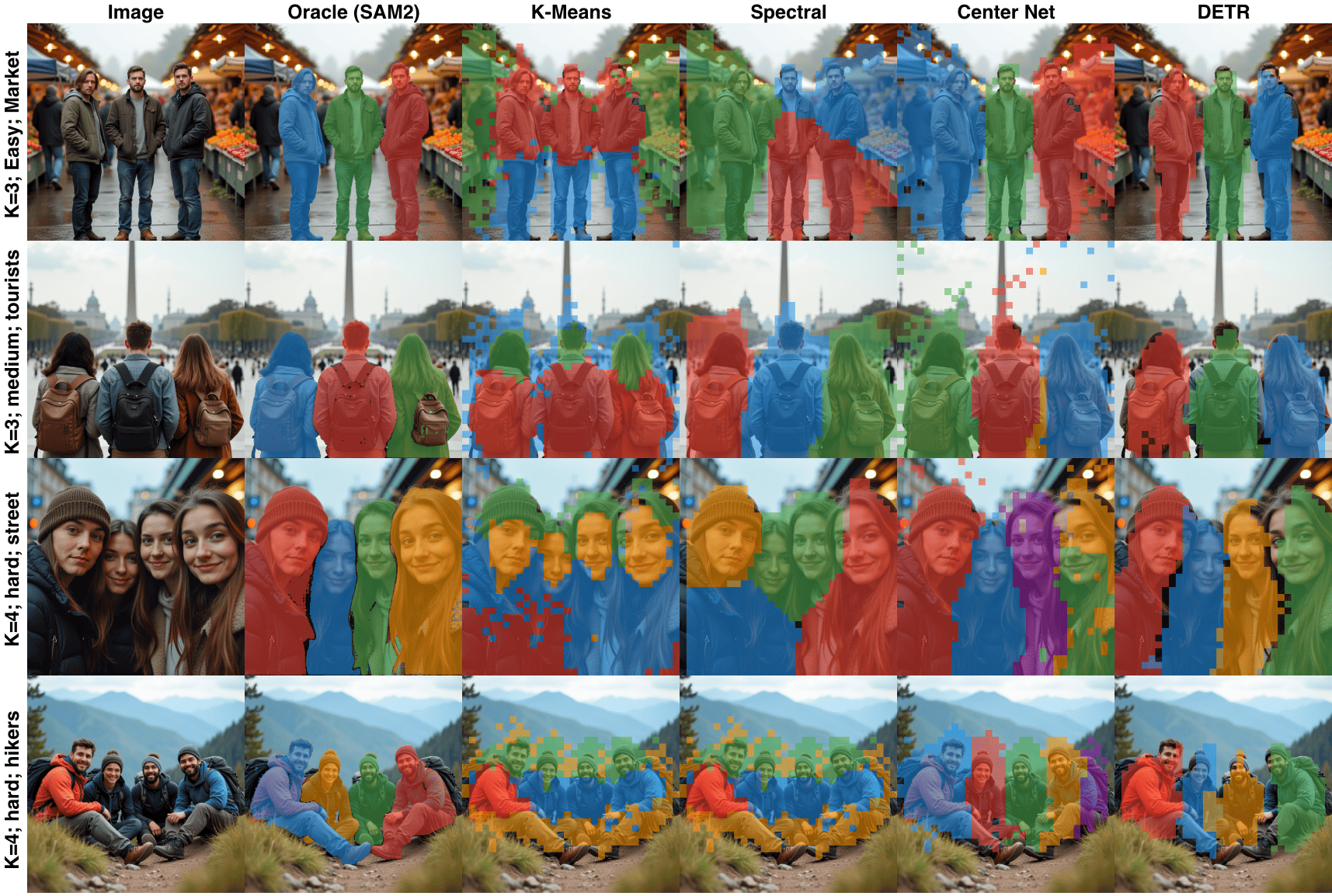}
    \caption{Qualitative comparison of instance segmentation methods across four scenes of increasing difficulty. 
    Each row shows the generated image, SAM2 oracle masks, and predictions from K-Means, Spectral clustering, CenterNet, and our DETR-based segmentor. 
    K-Means and Spectral produce fragmented, noisy masks in complex scenes (\textit{rows 3--4}), while CenterNet sometimes over-predicts the instance count under occlusion. 
    DETR consistently recovers the correct number of instances with coherent boundaries across all difficulty levels.}
    \label{app:qualitative_seg}
\end{figure*}

\section{Steering Performance by Person Count and Oracle-Mask Validation}
\label{app:person_count_oracle}
Figure~\ref{app:seg_difficulty_k} above shows that segmentor \emph{detection} quality (mIoU/AP) degrades with $K$; here we ask the downstream question directly: does that degradation propagate into \emph{steering} accuracy, and how much of any remaining error is attributable to the segmentor rather than the steering mechanism. Both experiments below use FLUX.1-Dev only, at the operating point $\alpha{=}2.0$, $\lambda{=}4.0$, window $[4,10)$.

\paragraph{Steering accuracy by $K$, at the operating point.}
Table~\ref{tab:steering_by_k} slices the steering-window ablation (Section~\ref{app:steering_hyperparameter}) by person count at $t_{\mathrm{start}}{=}4$, $t_{\mathrm{end}}{=}10$. This ablation's evaluation set covers $k\in\{2,3\}$ only, the same scope as the original steering-hyperparameter ablation it is drawn from; it does not include $k{=}4$.

\begin{table}[h]
\centering
\small
\caption{\textbf{Steering accuracy by person count} at the selected operating point ($n{=}80$/K).}
\label{tab:steering_by_k}
\begin{tabular}{lcccc}
\toprule
$K$ & $\mathrm{Acc}_{\mathrm{race}}$ & $\mathrm{Acc}_{\mathrm{gender}}$ & $K_{\mathrm{detected}}$ (mean) & $H_{ix}$ \\
\midrule
2 & 0.733 & 0.748 & 2.050 & 0.662 \\
3 & 0.683 & 0.766 & 3.075 & 0.861 \\
\bottomrule
\end{tabular}
\end{table}

\paragraph{Oracle-mask (SAM2) vs.\ DETR segmentor, matched pairs.}
To isolate the segmentor's contribution specifically, we re-run SGS with SAM2-derived masks (extracted from the finished, unsteered baseline image) in place of the DETR segmentor, on the same 445 images for both (identical occupation/$K$/template/seed; only the mask source differs).

\begin{table}[h]
\centering
\small
\caption{\textbf{DETR vs.\ oracle (SAM2) masks}, matched pairs, $n{=}445$ overall (149/148/148 at $K{=}2/3/4$).}
\label{tab:oracle_vs_detr}
\resizebox{\linewidth}{!}{%
\begin{tabular}{lccccc}
\toprule
$K$ & Mask source & $\mathrm{Acc}_{\mathrm{race}}$ & $\mathrm{Acc}_{\mathrm{gender}}$ & $H_{ix}$ & $\mathrm{LPIPS}_{\mathrm{bg}}$ \\
\midrule
\multirow{2}{*}{2} & DETR   & 0.843 & 0.896 & 0.792 & 0.743 \\
                    & Oracle & 0.812 & 0.849 & 0.799 & 0.750 \\
\multirow{2}{*}{3} & DETR   & 0.722 & 0.806 & 0.835 & 0.738 \\
                    & Oracle & 0.705 & 0.776 & 0.902 & 0.740 \\
\multirow{2}{*}{4} & DETR   & 0.643 & 0.720 & 0.834 & 0.756 \\
                    & Oracle & 0.701 & 0.764 & 0.874 & 0.753 \\
\midrule
\multirow{2}{*}{Overall} & DETR   & 0.714 & 0.808 & 0.820 & 0.746 \\
                          & Oracle & 0.740 & 0.796 & 0.858 & 0.748 \\
\bottomrule
\end{tabular}%
}
\end{table}

The DETR segmentor matches or exceeds oracle-mask race accuracy at $K{=}2$ and $K{=}3$ (0.843 vs.\ 0.812; 0.722 vs.\ 0.705), but falls behind it at $K{=}4$ (0.643 vs.\ 0.701, a gap of 0.058).
The Oracle SAM2 masks recover the race-steering accuracy that DETR loses once scenes become more crowded.
Gender accuracy shows a related but smaller pattern: DETR leads at $K=2$ and $K=3$  (0.896 vs.\ 0.849; 0.806 vs.\ 0.776) while the oracle mask lead at $K=4$ by 0.04 (0.764 vs.\ 0.720).
Background preservation ($\mathrm{LPIPS}_{\mathrm{bg}}$) is flat across mask source at every $K$, so this $K{=}4$ gap is not a fidelity tradeoff.
Taken together, these results indicate the live segmentor is not the limiting factor behind SGS's residual steering error at $K{=}2$ and $K{=}3$.
The gap to perfect accuracy comes from elsewhere in the pipeline (steering strength, classifier noise).

\section{Classifier Validation}
\label{app:classifier_validation}
All race/gender numbers in this paper depend on one automatic tool, FairFace~\cite{karkkainen2019fairface}. 
To read demographics off a generated face, before trusting it, we check how accurate it actually is against a set of images where the true demographic is known for certain, and we also check two other classifiers (a CLIP~\cite{radford2021learning} zero-shot classifier, and DeepFace~\cite{serengil2026boosted}) for comparison.

\paragraph{Building a set of images with known demographics.}
We generate images where every person in the group is asked for the same demographic (e.g.\ ``a group of three Black women''), so the prompt itself has a cue for every face in that image. 
This is done for both FLUX.1-Dev and SD3.5-Large, across all 8 race$\times$gender combinations and group sizes $k\in\{2,3,4\}$. We keep only images where a face detector actually finds exactly $k$ faces, to avoid mixing in blank or wrong-count images; this leaves 1{,}832 individual face crops with a known-correct label.

\paragraph{Results.}
Table~\ref{tab:classifier_validation} reports accuracy for each classifier on this set. FairFace and CLIP are both highly accurate on race (95--96\%) and near-perfect on gender ($>$99\%); DeepFace is noticeably weaker on race (88\%) and gender (93\%). ``Abstain'' means the classifier's raw output fell into a category we do not use (e.g.\ Middle Eastern, Latino/Hispanic, dropped due to high confusion with other categories reported in several previous works \cite{EA,shenfinetuning}); this happens occasionally for FairFace and DeepFace and never for CLIP, which only ever outputs one of our four race categories directly.

\begin{table}[h]
\centering
\small
\caption{\textbf{Classifier validation} against known-correct labels ($n{=}1{,}832$ face crops, both FLUX.1-Dev and SD3.5-Large).}
\label{tab:classifier_validation}
\begin{tabular}{lcccc}
\toprule
 & \multicolumn{2}{c}{Race} & \multicolumn{2}{c}{Gender} \\
\cmidrule(lr){2-3} \cmidrule(lr){4-5}
Classifier & Accuracy & Abstain & Accuracy & Abstain \\
\midrule
FairFace & 95.7\% & 2.5\% & 99.7\% & 0.0\% \\
CLIP     & 95.3\% & 0.0\% & 99.9\% & 0.0\% \\
DeepFace & 87.7\% & 7.5\% & 92.7\% & 0.0\% \\
\bottomrule
\end{tabular}
\end{table}

\paragraph{Do the three classifiers agree with each other?}
Separately from whether each classifier is correct, we check how often the three classifiers agree with each other (Krippendorff's alpha, where 1.0 means perfect agreement and 0 means no better than chance). Agreement is high on race (0.94) and on gender (0.94), and every pairwise comparison between two classifiers also shows strong agreement (Cohen's kappa between 0.85 and 1.00 in every case). This gives an independent check on the results above: it is not just that FairFace happens to agree with the ground-truth labels we built, three separately-built tools broadly agree with each other too.

\section{Segmentor Timestep Sweep -- SD3.5-Large}
\label{app:segmentor_sweep_sd3}
We repeat the segmentor timestep sweep of Section~\ref{sec:seg_results} on SD3.5-Large, training a separate DETR-style segmentor on SD3.5's own self-attention features at $1024{\times}1024$ resolution (same architecture and training recipe as the FLUX segmentor, Section~\ref{app:segmentor}), and evaluating on a held-out validation split ($n{=}288$) of SD3.5-generated images. Across SD3.5's 28-step denoising schedule, segmentation quality peaks at timestep index 17, reaching mIoU of \textbf{0.664}, AP50 of \textbf{0.841}, and AP50-95 of \textbf{0.413} which is comparable to FLUX's own peak (mIoU 0.664, AP50 0.797, AP50-95 0.416, Section~\ref{sec:seg_results}), indicating that MM-DiT self-attention features become instance-discriminative at an analogous point in SD3.5's denoising trajectory as well.

\begin{figure}[h]
    \centering
    \includegraphics[width=\linewidth]{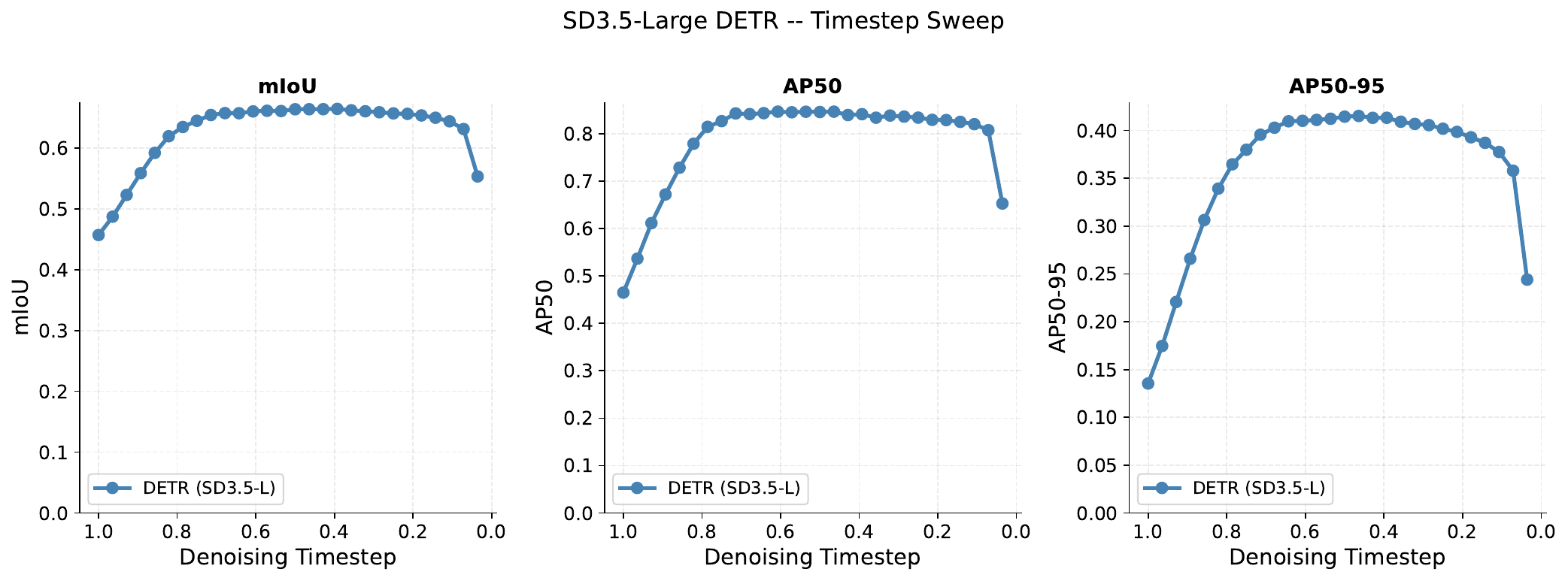}
    \caption{\textbf{SD3.5-Large DETR segmentor, timestep sweep} (val split, $n{=}288$). mIoU, AP50, and AP50-95 across SD3.5's 28-step denoising schedule, mirroring Figure~\ref{fig:arch_sweep}'s FLUX sweep. Performance peaks at timestep index 17 and plateaus around it, the same qualitative shape as FLUX's own sweep.}
    \label{app:sd3_sweep_fig}
\end{figure}

\section{Segmentor Results - Qualitative}
\label{app:segmentor_qualitative_results}
Figure \ref{app:qualitative_seg} compares instance segmentation predictions across four scenes spanning easy to hard difficulty.
In the simpler market and tourist scenes (\textit{rows 1--2}), all methods recover the correct instance count, though K-Means and Spectral clustering exhibit scattered noise pixels caused by high-frequency variation in the attention feature space.
The gap widens in the close-up street scene and the outdoor hikers scene (\textit{rows 3--4}): K-Means and Spectral produce heavily fragmented masks that merge or split individuals, while CenterNet over-estimates the instance count due to ambiguous overlap regions.
DETR-based segmentor produces coherent per-person masks with the correct instance counts, demonstrating that the cross-attention mechanism of DETR learns to separate spatially proximate instances that confound clustering-based approaches.

\begin{table}[h]
\centering
\small
\caption{
    \textbf{Steering hyperparameter ablation.} \textbf{Bold} = selected configuration.
    $\mathrm{Aes}\Delta$: aesthetic score delta vs.\
    unsteered baseline (higher $=$ better quality).
    $\mathrm{LPIPS}_{\mathrm{bg}}$: background preservation.
    $\mathrm{CLIP}\Delta$: prompt--image CLIP similarity delta
    (closer to 0 $=$ less semantic drift).
}
\label{tab:steering_ablation_supp}
\resizebox{\linewidth}{!}{%
\begin{tabular}{cccccc}
    \toprule
    & $\mathrm{Acc}_{\mathrm{race}}$$\uparrow$
    & $\mathrm{Acc}_{\mathrm{gender}}$$\uparrow$
    & $\mathrm{Aes}\Delta$$\uparrow$
    & $\mathrm{LPIPS}_{\mathrm{bg}}$$\uparrow$
    & $\mathrm{CLIP}\Delta$$(\rightarrow 0)$ \\
    \midrule
    \multicolumn{6}{l}{%
        \textit{(a) EA strength $\alpha$
        \normalfont{[$\lambda{=}4.0$ fixed]}}} \\[3pt]
    1.5          & 0.573 & 0.722 & +0.009 & 0.916 & $-$0.026 \\
    \textbf{2.0} & \textbf{0.708} & \textbf{0.757}
                 & \textbf{+0.027} & \textbf{0.900}
                 & \textbf{$-$0.033} \\
    2.5          & 0.723 & 0.740 & +0.010 & 0.890 & $-$0.042 \\
    3.0          & 0.636 & 0.738 & $-$0.134 & 0.876 & $-$0.052 \\
    3.5          & 0.438 & 0.566 & $-$0.762 & 0.842 & $-$0.104 \\
    \midrule
    \multicolumn{6}{l}{%
        \textit{(b) Spatial strength $\lambda$
        \normalfont{[$\alpha{=}2.0$ fixed]}}} \\[3pt]
    3.0          & 0.644 & 0.734 & +0.020 & 0.914 & $-$0.028 \\
    \textbf{4.0} & \textbf{0.708} & \textbf{0.757}
                 & \textbf{+0.027} & \textbf{0.900}
                 & \textbf{$-$0.033} \\
    5.0          & 0.736 & 0.742 & +0.034 & 0.889 & $-$0.039 \\
    6.0          & 0.767 & 0.746 & $-$0.019 & 0.876 & $-$0.046 \\
    7.0          & 0.758 & 0.742 & $-$0.120 & 0.864 & $-$0.053 \\
    \bottomrule
\end{tabular}}
\end{table}

\section{Steering Hyperparameter Sensitivity}
\label{app:steering_hyperparameter}

Table~\ref{tab:steering_ablation_supp} reports the full sensitivity analysis for EA strength $\alpha$ and spatial diversity strength $\lambda$, using the evaluation setup described in Section~5.4 of the main paper.

For $\alpha$ (panel a), steering accuracy and aesthetic quality improve from $\alpha{=}1.5$ to $\alpha{=}2.0$, then degrade at higher values. $\alpha{=}2.5$ achieves marginally higher race accuracy (0.723 vs.\ 0.708) but at the cost of gender accuracy, aesthetic quality, background preservation, and semantic coherence.
At $\alpha{\geq}3.0$, aesthetic quality falls below the unsteered baseline and CLIP similarity drops sharply, indicating the shifted embedding has diverged from the original prompt semantics.

For $\lambda$ (panel b), $\lambda{=}4.0$ achieves the best joint performance. 
Higher values improve race accuracy marginally but incur increasing aesthetic degradation and semantic drift; $\lambda{=}7.0$ produces a negative aesthetic delta ($-$0.120) and a CLIP drop of $-$0.053. 
We select $\alpha{=}2.0$ and $\lambda{=}4.0$ as our operating point.

\begin{table}[h]
\centering
\small
\caption{
    \textbf{Steering window placement ablation} (FLUX.1-Dev, $\alpha{=}2.0$, $\lambda{=}4.0$).
    \textbf{Bold} = selected operating point.
    CLIP$\Delta$: prompt--image similarity delta vs.\ baseline.
}
\label{tab:window_ablation_supp}
\resizebox{\linewidth}{!}{%
\setlength{\tabcolsep}{5pt}
\begin{tabular}{clccccc}
    \toprule
    $Start$ & $End$
    & $\mathrm{Acc}_{\mathrm{race}}$$\uparrow$
    & $\mathrm{Acc}_{\mathrm{gender}}$$\uparrow$
    & $\mathrm{Aes}\Delta$$\uparrow$
    & $\mathrm{LPIPS}_{\mathrm{bg}}$$\uparrow$
    & $\mathrm{CLIP}\Delta$$\rightarrow 0$ \\
    \midrule
    \multirow{4}{*}{2}
        & 8  & 0.713 & 0.784 & $-$0.097 & 0.760 & $-$0.097 \\
        & 10 & 0.718 & 0.783 & $-$0.123 & 0.759 & $-$0.102 \\
        & 12 & 0.749 & 0.791 & $-$0.179 & 0.758 & $-$0.105 \\
        & 14 & 0.754 & 0.783 & $-$0.237 & 0.757 & $-$0.106 \\
    \midrule
    \multirow{4}{*}{4}
        & 8        & 0.632 & 0.744 & +0.008 & 0.904 & $-$0.029 \\
        & \textbf{10} & \textbf{0.708} & \textbf{0.757}
                   & \textbf{+0.027} & \textbf{0.900} & \textbf{$-$0.033} \\
        & 12       & 0.741 & 0.743 & +0.036 & 0.899 & $-$0.038 \\
        & 14       & 0.769 & 0.745 & +0.036 & 0.898 & $-$0.041 \\
    \midrule
    \multirow{3}{*}{6}
        & 10 & 0.476 & 0.649 & $-$0.001 & 0.961 & $-$0.013 \\
        & 12 & 0.545 & 0.652 & +0.002   & 0.958 & $-$0.016 \\
        & 14 & 0.594 & 0.649 & +0.007   & 0.956 & $-$0.018 \\
    \bottomrule
\end{tabular}%
}
\end{table}

\paragraph{Steering window}
Early detection and steering at $t_{\mathrm{start}}{=}2$ yields the highest steering accuracy across all end points but produces consistently negative aesthetic deltas and poor background preservation (LPIPS$_{\mathrm{bg}}{\leq}0.759$), indicating that MM-DiT features at step 2 are insufficiently structured for reliable instance segmentation, causing the steering signal to leak beyond the intended mask regions.
Late detection and steering at $t_{\mathrm{start}}{=}{6}$ preserves background coherence well (LPIPS$_{\mathrm{bg}}{\geq}0.956$) but achieves substantially lower steering accuracy ($\mathrm{Acc}_{\mathrm{race}}{\leq}0.594$, $\mathrm{Acc}_{\mathrm{gender}}{\leq}0.652$), as the remaining steering window is too short to shift demographics effectively.
$t_{\mathrm{start}}{=}4$ achieves the best balance across all metrics: extending the end step from 8 to 10 improves steering accuracy (0.632$\to$0.708 race, 0.744$\to$0.757 gender) with only marginal reductions in background preservation, while further extension to $t_{\mathrm{end}}{=}12$ or $14$ offers diminishing accuracy gains at the cost of increasing semantic drift.
We therefore select $t_{\mathrm{start}}{=}4$ and $t_{\mathrm{end}}{=}10$ as our operating point, corresponding to steps 4--10 of the 21-step denoising trajectory.

\section{Steering Hyperparameter and Window Ablation -- SD3.5-Large}
\label{app:steering_hyperparameter_sd3}

We repeat the steering-strength and steering-window ablations on SD3.5-Large.

\begin{table}[h]
\centering
\small
\caption{\textbf{SD3.5-Large steering-strength ablation.} $\mathrm{LPIPS}_{\mathrm{bg}}$: background preservation.}
\label{tab:sd3_strength_ablation_supp}
\begin{tabular}{ccccc}
\toprule
$\alpha$ & $\lambda$ & $\mathrm{Acc}_{\mathrm{race}}$$\uparrow$ & $\mathrm{Acc}_{\mathrm{gender}}$$\uparrow$ & $\mathrm{LPIPS}_{\mathrm{bg}}$$\uparrow$ \\
\midrule
1.0 & 1.0 & 0.443 & 0.620 & 0.958 \\
1.0 & 2.0 & 0.526 & 0.646 & 0.938 \\
1.0 & 4.0 & 0.707 & 0.693 & 0.904 \\
1.5 & 1.0 & 0.561 & 0.653 & 0.946 \\
1.5 & 2.0 & 0.694 & 0.677 & 0.912 \\
1.5 & 4.0 & 0.809 & 0.722 & 0.877 \\
2.0 & 1.0 & 0.658 & 0.660 & 0.931 \\
2.0 & 2.0 & 0.752 & 0.682 & 0.893 \\
\textbf{2.0} & \textbf{4.0} & \textbf{0.800} & \textbf{0.759} & \textbf{0.855} \\
\bottomrule
\end{tabular}%
\end{table}

Table~\ref{tab:sd3_strength_ablation_supp} sweeps $\alpha\in\{1.0,1.5,2.0\}$ and $\lambda\in\{1.0,2.0,4.0\}$. Unlike FLUX, accuracy improves monotonically across the tested range on both parameters, with $\alpha{=}2.0$, $\lambda{=}4.0$ the best-performing configuration tested; we select this as SD3.5's operating point.

\begin{table}[h]
\centering
\small
\caption{\textbf{SD3.5-Large steering-window ablation}, $\alpha{=}2.0$, $\lambda{=}4.0$.}
\label{tab:sd3_window_ablation_supp}
\begin{tabular}{clccc}
\toprule
$Start$ & $End$ & $\mathrm{Acc}_{\mathrm{race}}$ & $\mathrm{Acc}_{\mathrm{gender}}$ & $\mathrm{LPIPS}_{\mathrm{bg}}$ \\
\midrule
\multirow{3}{*}{3}
    & 10 & 0.391 & 0.624 & 0.921 \\
    & 12 & 0.402 & 0.624 & 0.920 \\
    & 14 & 0.427 & 0.642 & 0.919 \\
\midrule
\multirow{5}{*}{4}
    & 11 & 0.378 & 0.623 & 0.945 \\
    & 13 & 0.397 & 0.617 & 0.944 \\
    & 16 & 0.424 & 0.617 & 0.944 \\
    & 19 & 0.453 & 0.626 & 0.944 \\
    & \textbf{24} & \textbf{0.514} & 0.622 & 0.944 \\
\midrule
\multirow{3}{*}{5}
    & 10 & 0.338 & 0.610 & 0.964 \\
    & 12 & 0.357 & 0.613 & 0.963 \\
    & 14 & 0.386 & 0.613 & 0.962 \\
\midrule
\multirow{3}{*}{6}
    & 10 & 0.293 & 0.590 & 0.977 \\
    & 12 & 0.310 & 0.590 & 0.976 \\
    & 14 & 0.340 & 0.595 & 0.975 \\
\bottomrule
\end{tabular}%
\end{table}

Table~\ref{tab:sd3_window_ablation_supp} sweeps detection/steering start step $\in\{3,4,5,6\}$ and end step, out of SD3.5's 28-step schedule. $\mathrm{start}{=}4$, $\mathrm{end}{=}24$ achieves the highest accuracy among the configurations tested.

\begin{table*}[t]
\centering
\small
\caption{
    \textbf{Per-occupation demographic metrics, baseline vs.\ SGS} (FLUX.1-Dev, extended benchmark, $n{=}150$/occupation).
    $H_{ix}$: intra-image intersectional entropy (higher = more diverse). $DR_{\mathrm{race}}$, $DR_{\mathrm{gender}}$: deviation ratio from uniform (lower = more balanced).
    Left block: male-skewed occupations. Right block: female-skewed occupations.
}
\label{tab:per_occupation_flux}
\begin{minipage}[t]{0.49\linewidth}
\centering
\resizebox{\linewidth}{!}{%
\begin{tabular}{lcccccc}
\toprule
& \multicolumn{3}{c}{Baseline} & \multicolumn{3}{c}{SGS} \\
\cmidrule(lr){2-4} \cmidrule(lr){5-7}
Occupation & $H_{ix}$ & $DR_r$ & $DR_g$ & $H_{ix}$ & $DR_r$ & $DR_g$ \\
\midrule
Driver      & 0.480 & 0.544 & 1.000 & 0.578 & 0.324 & 0.680 \\
Mover       & 0.406 & 0.486 & 0.920 & 0.717 & 0.297 & 0.627 \\
Developer   & 0.482 & 0.794 & 0.987 & 0.843 & 0.279 & 0.787 \\
Farmer      & 0.426 & 0.759 & 0.987 & 0.761 & 0.324 & 0.587 \\
Guard       & 0.503 & 0.333 & 0.813 & 0.661 & 0.270 & 0.720 \\
Chief       & 0.505 & 0.116 & 0.707 & 0.859 & 0.315 & 0.267 \\
Janitor     & 0.547 & 0.674 & 0.800 & 0.810 & 0.306 & 0.600 \\
Lawyer      & 0.582 & 0.866 & 0.840 & 0.735 & 0.289 & 0.880 \\
Cook        & 0.489 & 0.687 & 0.893 & 0.858 & 0.421 & 0.267 \\
Physician   & 0.598 & 0.640 & 0.640 & 0.862 & 0.280 & 0.587 \\
CEO         & 0.385 & 0.907 & 0.987 & 0.793 & 0.288 & 0.987 \\
Analyst     & 0.563 & 0.850 & 1.000 & 0.852 & 0.297 & 0.947 \\
Manager     & 0.469 & 0.916 & 0.987 & 0.859 & 0.280 & 0.840 \\
Supervisor  & 0.472 & 0.935 & 0.973 & 0.852 & 0.289 & 0.867 \\
Salesperson & 0.748 & 0.917 & 0.733 & 0.888 & 0.387 & 0.627 \\
\bottomrule
\end{tabular}%
}
\end{minipage}
\hfill
\begin{minipage}[t]{0.49\linewidth}
\centering
\resizebox{\linewidth}{!}{%
\begin{tabular}{lcccccc}
\toprule
& \multicolumn{3}{c}{Baseline} & \multicolumn{3}{c}{SGS} \\
\cmidrule(lr){2-4} \cmidrule(lr){5-7}
Occupation & $H_{ix}$ & $DR_r$ & $DR_g$ & $H_{ix}$ & $DR_r$ & $DR_g$ \\
\midrule
Editor      & 0.605 & 0.872 & 0.573 & 0.772 & 0.298 & 0.387 \\
Designer    & 0.603 & 0.732 & 0.147 & 0.777 & 0.333 & 0.133 \\
Accountant  & 0.642 & 0.927 & 0.813 & 0.869 & 0.374 & 0.853 \\
Auditor     & 0.492 & 0.816 & 0.973 & 0.758 & 0.324 & 0.813 \\
Writer      & 0.622 & 0.813 & 0.440 & 0.826 & 0.306 & 0.200 \\
Baker       & 0.420 & 0.621 & 0.787 & 0.806 & 0.298 & 0.280 \\
Clerk       & 0.534 & 0.890 & 0.947 & 0.857 & 0.289 & 0.760 \\
Cashier     & 0.514 & 0.854 & 0.440 & 0.783 & 0.270 & 0.040 \\
Counselor   & 0.623 & 0.937 & 0.080 & 0.824 & 0.279 & 0.200 \\
Attendant   & 0.686 & 0.841 & 0.413 & 0.922 & 0.289 & 0.347 \\
Tailor      & 0.395 & 0.756 & 0.973 & 0.861 & 0.324 & 0.653 \\
Librarian   & 0.515 & 0.776 & 0.467 & 0.722 & 0.494 & 0.040 \\
Assistant   & 0.566 & 0.757 & 0.080 & 0.805 & 0.324 & 0.107 \\
Cleaner     & 0.507 & 0.569 & 0.360 & 0.650 & 0.297 & 0.520 \\
Hairdresser & 0.355 & 0.911 & 0.547 & 0.667 & 0.391 & 0.333 \\
\bottomrule
\end{tabular}%
}
\end{minipage}
\end{table*}

\section{Extended Benchmark -- Occupations and Prompt Templates}
\label{app:extended_benchmark_protocol}
Table~\ref{tab:benchmark} in the main paper (Section~\ref{sec:benchmark}) evaluates on 30 WinoBias occupations~\cite{zhao2018gender}, disjoint from the 10 occupations used to fit the EA direction vectors (\textit{carpenter, mechanic, construction worker, laborer, sheriff, teacher, nurse, housekeeper, flight attendant, librarian}), so all baselines are evaluated out-of-distribution with respect to their own calibration set.

\paragraph{Occupations.}
15 male-skewed: \textit{driver, mover, developer, farmer, guard, chief, janitor, lawyer, cook, physician, CEO, analyst, manager, supervisor, salesperson}.
15 female-skewed: \textit{editor, designer, accountant, auditor, writer, baker, clerk, cashier, counselor, attendant, tailor, librarian, assistant, cleaner, hairdresser}.

\paragraph{Prompt templates.}
Each occupation is instantiated under 5 templates, with $\{k\}$ and \{occupation\} substituted as in the main paper:
\begin{enumerate}
    \item \textit{"A portrait photo of $\{k\}$ \{occupation\}s"}
    \item \textit{"A photo of $\{k\}$ \{occupation\}s"}
    \item \textit{"A group of $\{k\}$ \{occupation\}s"}
    \item \textit{"A realistic image of $\{k\}$ \{occupation\}s at work"}
    \item \textit{"$\{k\}$ \{occupation\}s standing together"}
\end{enumerate}
Combined with $k\in\{2,3,4\}$ and 10 seeds, this yields $30 \times 5 \times 3 \times 10 = 4{,}500$ images per method.

\section{Per-Occupation Demographic Distribution}
Table~\ref{tab:per_occupation_flux} breaks down intra-image intersectional entropy ($H_{ix}$) and aggregate deviation ratio ($DR_{\mathrm{race}}$, $DR_{\mathrm{gender}}$) per occupation, for the baseline and SGS, across all 30 occupations of the extended WinoBias benchmark (Section~5.5 of the main paper; FLUX.1-Dev). Occupations are grouped by the male-skewed/female-skewed split used to construct the benchmark (Section~\ref{app:extended_benchmark_protocol}).

\paragraph{Entropy.}
SGS raises $H_{ix}$ over the baseline for every one of the 30 occupations without exception -- the smallest gain is Driver (0.480$\to$0.578) and the largest is Tailor (0.395$\to$0.861), with no occupation where SGS's per-instance steering fails to increase intra-image diversity relative to the unsteered model.

\paragraph{Aggregate deviation.}
$DR_{\mathrm{race}}$ drops for every occupation under SGS, typically substantially (e.g.\ Manager: 0.916$\to$0.280; CEO: 0.907$\to$0.288), confirming the aggregate race-balance improvement seen in Table~\ref{tab:benchmark} holds occupation-by-occupation, not just on average. $DR_{\mathrm{gender}}$ is noisier at the per-occupation level and does not improve uniformly -- most occupations move closer to balance (e.g.\ Chief: 0.707$\to$0.267; Cashier: 0.440$\to$0.040) but five of the 30 move further from it (Lawyer: 0.840$\to$0.880; Accountant: 0.813$\to$0.853; Counselor: 0.080$\to$0.200; Assistant: 0.080$\to$0.107; Cleaner: 0.360$\to$0.520) -- consistent with gender assignment being a coarser 2-category balancing problem where per-occupation sample size ($n{=}150$) leaves more room for variance than the pooled Table~\ref{tab:benchmark} figures show.

\paragraph{Intersectional Observation.}
Taken together, Table~\ref{tab:per_occupation_flux} confirms the core motivation for $H_{ix}$ holds at the level of individual occupations, not only in aggregate: the baseline is both gender-stereotyped and racially homogeneous on essentially every occupation tested, and SGS's per-instance steering raises intra-image diversity consistently across all of them.

\section{Extended Qualitative Analysis}
\label{app:supp_qualitative_sec}
\begin{figure*}[h]
    \centering
    \includegraphics[width=\linewidth]
        {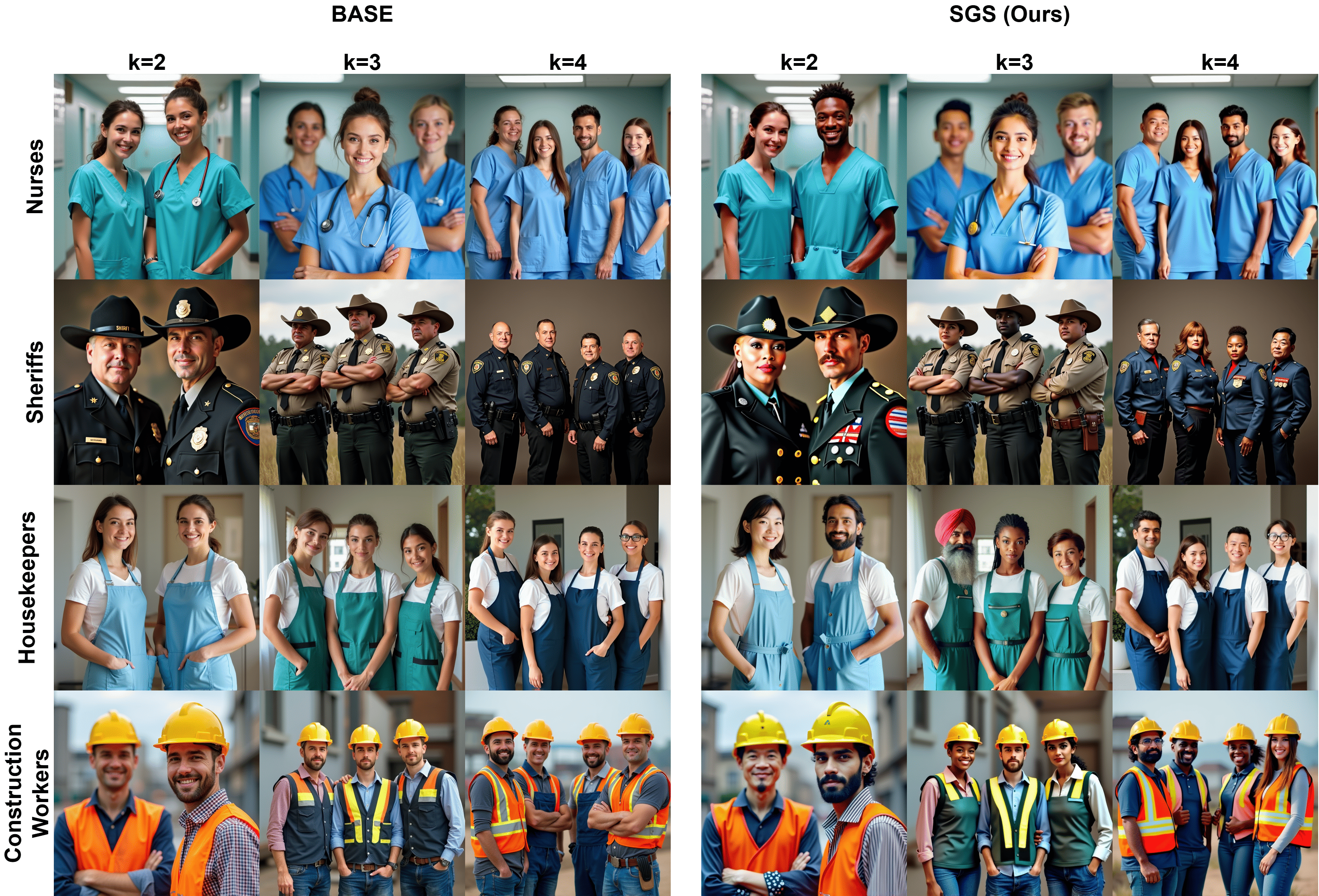}
    \caption{
        \textbf{Extended qualitative comparison} across four occupations and three person counts $k \in \{2, 3, 4\}$.
        \emph{Left}: baseline FLUX outputs.
        \emph{Right}: SGS outputs.
        Rows show female-dominant (Nurses, Housekeepers) and male-dominant (Sheriffs, Construction Workers) professions.
        Baseline outputs are demographically homogeneous across all $k$ values and both stereotype directions.
        SGS consistently produces intersectional demographic diversity --- each person within the same scene is steered to a distinct demographic target --- while also preserving occupational context (uniform, apron, badge, hard hat) and scene coherence across all conditions.
        Performance is consistent across $k \in \{2,3,4\}$, demonstrating that the segmentor reliably separates and independently steers two, three, and four persons within the same image.
    }
    \label{app:supp_qualitative}
\end{figure*}

Figure~\ref{app:supp_qualitative} presents an extended qualitative comparison across four occupations spanning both stereotype directions and three person counts $k \in \{2,3,4\}$.
Occupations include two female-dominated professions (Nurse, Housekeeping) and two male-dominated professions (Sheriffs, Construction Workers).

\paragraph{Baseline.}
Across all four occupations and all values of $k$, the unsteered baseline produces demographically homogeneous outputs that reflect occupational stereotypes;
Nurses and Housekeepers are consistently female-presenting or White, while Sheriffs and Construction workers are consistently male-presenting and White.
This homogeneity persists regardless of $k$, confirming that the stereotyping is driven by the text prompt rather than any scene composition effect introduced by specifying multiple persons.

\paragraph{SGS across person counts.}
SGS consistently produces intersectional demographic diversity across all $k$ values.
At $k=2$, each pair contains two persons steered to distinct intersectional targets-- for example, a Black male nurse is paired with an Asian female nurse, or an Indian male housekeeper paired with an Asian female housekeeper.
At $k=3$ and $k=4$, the segmentor reliably separates all persons and assigns independent demographic vectors to each, with the resulting groups spanning multiple racial categories and both genders within the same image.
The consistency across $k$ values demonstrates that the instance segmentor remains reliable as scene complexity increases.

\paragraph{Occupational context preservation.}
A key requirement of practical fairness steering is that demographic edits do not disrupt occupational cues.
Across all conditions, SGS preserves the visual markers of each profession: nurses retain scrubs and stethoscopes, sheriffs retain badges and hats, housekeepers retain aprons, and construction workers retain safety helmets and reflective vests.
This confirms that the spatially gated steering is confined to person regions via the instance masks, preventing demographic steering from leaking into occupational attire or scene background.

\section{Qualitative Comparison -- SD3.5-Large}
\label{app:supp_qualitative_sd3}
\begin{figure*}[t]
    \centering
    \includegraphics[width=\linewidth]{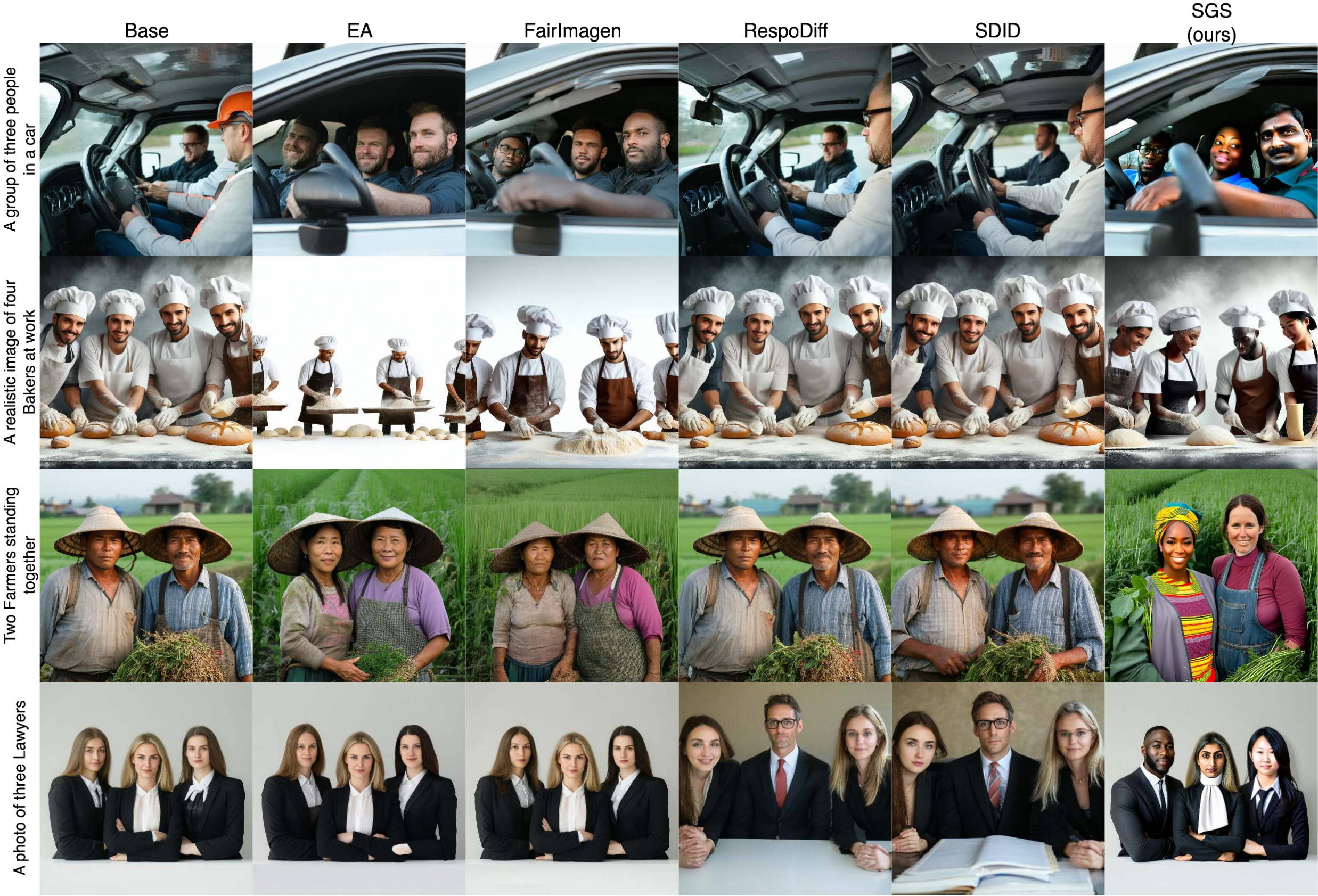}
    \caption{\textbf{Qualitative comparison on SD3.5-Large}, mirroring Figure~\ref{fig:qualitative}. Base, EA, FairImagen, RespoDiff, and SDID remain homogeneous or only mildly shifted across all four prompts; SGS independently diversifies each detected person while preserving occupational context.}
    \label{app:supp_qualitative_sd3_fig}
\end{figure*}

Figure~\ref{app:supp_qualitative_sd3_fig} repeats the main paper's qualitative comparison (Figure~\ref{fig:qualitative}) on SD3.5-Large, across four prompts and all five baselines. 
Images from Base, EA, FairImagen, RespoDiff, and SDID remain demographically homogeneous or only mildly shifted across all four scenes, while SGS diversifies every scene independently per person while preserving occupational context (aprons, rice-field setting, formal attire).
This confirms the pattern shown quantitatively in Table~\ref{tab:benchmark} holds qualitatively on a second architecture as well.

\section{Baseline Adaptation to Diffusion Transformers}
\label{app:baseline_adaptation}
FairImagen~\citep{fu2026fairimagen}, SDID~\citep{li2024self}, and RespoDiff~\citep{respodiff} were each designed and released against older UNet-based diffusion models -- FairImagen against SD1.5 and SDXL, RespoDiff against SDXL, and SDID against a UNet-based Stable Diffusion pipeline -- where generation is driven by a single-stream convolutional UNet with cross-attention to a text encoder. FLUX and SD3.5-Large instead use a Multimodal Diffusion Transformer (MM-DiT, Section~\ref{sec:mmdit}): image and text tokens are processed jointly through a stack of self-attention blocks, with no UNet and no single bottleneck layer. None of the three methods' released implementations run on this architecture family as-is. We faithfully reimplemented each method's core mechanism for MM-DiT, verified directly against its own released code, rather than mechanically porting UNet-specific hooks.

\paragraph{FairImagen.}
The original method operates on the model's pooled text-conditioning vector. For each demographic group $g$, it defines an empirical bias direction
\begin{equation}
    \nu_g = \frac{1}{|X^{(g)}|}\sum_{\bar{E}_p \in X^{(g)}} \bar{E}_p - \bar{E},
\end{equation}
and obtains a projector $P$ by solving
\begin{equation}
    \min_{P^\top P = I} \; -\mathrm{Tr}(P^\top \Sigma_X P) + \lambda \lVert BP \rVert_F^2,
\end{equation}
where $B = Z^\top X \in \mathbb{R}^{G\times D}$ collects the per-group mean directions and $Z \in \{0,1\}^{n\times G}$ is a one-hot group indicator; $P$ thus spans the top-variance directions within the null space of the group-dependent signal $B$. At inference, the projector is applied orthogonally to the pooled embedding, $\bar{E}_p' = PP^\top\bar{E}_p$, before it conditions the UNet. MM-DiT models expose an analogous pooled conditioning vector alongside their token-level sequence conditioning, so we fit and apply this same projector on this real pooled vector, leaving the token sequence untouched.

\paragraph{SDID.}
The original method learns a single concept offset $c \in \mathbb{R}^D$ per demographic direction by
\begin{equation}
    c^* = \arg\min_c \sum_{x,y\sim\mathcal{D}}\sum_{t\sim[0,T]} \lVert \epsilon - \epsilon_\theta(x_t^+, t, \pi(y^-), c) \rVert^2,
\end{equation}
where $x^+$ is an image generated from a concept-inclusive prompt and $y^-$ is the same prompt with the concept term removed, with the base model $\epsilon_\theta$ kept frozen throughout. At inference, the offset is added directly to the UNet's image-latent bottleneck activation, $h \leftarrow h + c$, a single injection point on the image side of the network. MM-DiT has no UNet bottleneck to target, so we carry the same design principle over: the learned offset is added to the hidden state of one representative block within the joint-transformer stack, preserving a single image-side injection point rather than broadcasting the offset into the text-token stream.

\paragraph{RespoDiff.}
The original method decomposes its responsible transformation as
\begin{equation}
    \hat f(y_{\mathrm{neu}}) = g\bigl(T_\theta^{\mathrm{resp},s}(h_{\mathrm{neu}}) + T_\theta^{\mathrm{sem},s}(h_{\mathrm{neu}})\bigr),
\end{equation}
training its Responsible Concept Alignment Module $T_\theta^{\mathrm{resp},s}$ (our race-aware/gender-aware modules) and Semantic Alignment Module $T_\theta^{\mathrm{sem},s}$ with
\begin{equation}
    \mathcal{L}_{\mathrm{resp}} = \mathbb{E}_{z_{t,\mathrm{neu}}}\bigl[\lVert \epsilon_{f_{\mathrm{resp}}}(z_{t,\mathrm{neu}}, y_{\mathrm{neu}}) - \epsilon_f(z_{t,\mathrm{neu}}, y_{\mathrm{tar}}^s)\rVert_2^2\bigr],
\end{equation}
\begin{equation}
    \mathcal{L}_{\mathrm{sem}} = \mathbb{E}_{z_{t,\mathrm{neu}}}\bigl[\lVert \epsilon_{\hat f}(z_{t,\mathrm{neu}}, y_{\mathrm{neu}}) - \epsilon_f(z_{t,\mathrm{neu}}, y_{\mathrm{neu}})\rVert_2^2\bigr],
\end{equation}
and injects their combined residual once at the UNet's bottleneck hidden state $h$. Our first attempt at reimplementing this for MM-DiT broadcast the same residual across every block of the joint-transformer stack; this reliably broke generation, producing incoherent output regardless of prompt or seed. Injecting the residual at a single representative block instead resolved this and produced coherent, steerable output.
This also brings the reimplementation closer to the original method's own single-bottleneck design than the all-block version we started from.

\paragraph{Shared adaptations.}
All three methods are applied only to the conditional half of the classifier-free-guidance batch, never the unconditional branch, consistent with how each was originally deployed. 
Training uses each target architecture's own native noise parametrization and scheduler rather than a generic substitute.

\paragraph{Code Availability.}
Reference implementations of all three adapted baselines are included in the supplementary code (\texttt{baselines/flux/} and \texttt{baselines/sd35/}): \texttt{fairimagen.py}, \texttt{sdid.py}, and \texttt{respodiff.py}, each containing the fit/training function and manipulation function described above with inline comments matching this section.
SGS itself is in \texttt{sgs/} (FLUX) and \texttt{sgs\_sd3/} (SD3.5-Large).

\section{Human Evaluation Protocol}
\label{sec:supp-human-eval}
\begin{figure}[h]
    \centering
    \includegraphics[width=0.70\linewidth]{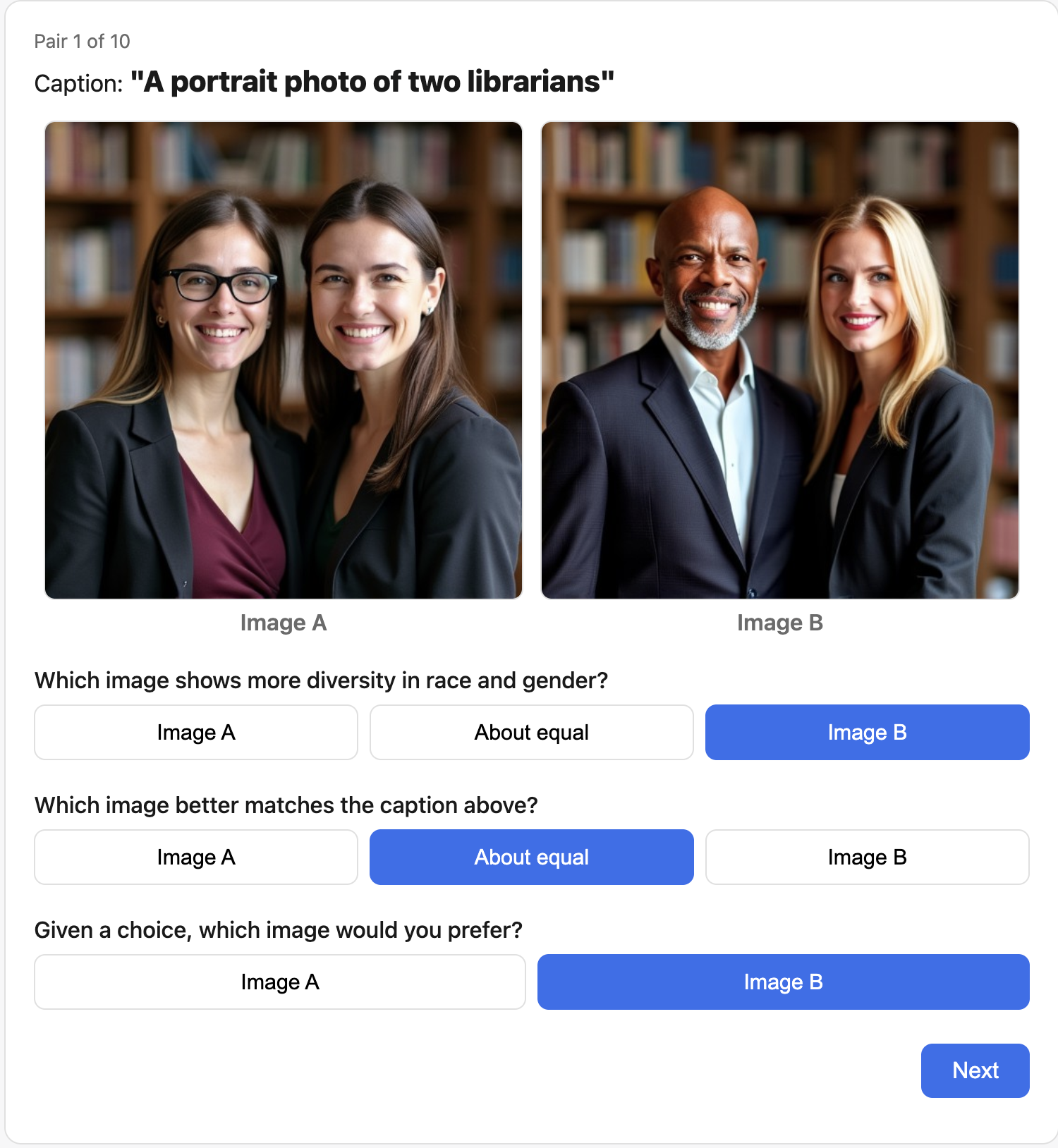}
    \caption{The human evaluation interface. Raters see a captioned imagepair with method identity and position blinded, and answer three questions per pair: perceived diversity, prompt faithfulness, and overall preference.}
    \label{fig:supp-survey-ui}
\end{figure}

Automatic demographic classifiers may not fully capture how a human observer perceives diversity, prompt fidelity and preference.
So we conducted a blind pairwise human evaluation comparing SGS against five baselines (Baseline, EA, FairImagen, RespoDiff and SDID) across 33 scenes spanning WinoBias occupations  at $K\in\{2,3,4\}$ people per image, using a fixed seed per scene shared across all six methods.
Each rater was shown two images side by side from the same scene, generated by two different methods (one always SGS), with method identity withheld (labeled only 'Image A'/'Image B') and left -- right placement randomized per pair.
For each pair, raters answered three questions (refer Figure~\ref{fig:supp-survey-ui}):
\begin{enumerate}
    \item Which image shows more diversity in race and gender?
    \item Which image better matches the caption above?
    \item Given a choice, which image would you prefer?
\end{enumerate}
The first two questions offered a three way choice (Image A / about equal / Image B), while the last was a forced binary choice used to obtain a holistic preference signal.
Each participants rated 10 pairs, two per comparison, ensuring balanced coverage across all five comparisons within a session.

Sessions in which a rater selected the identical response to every question throughout were excluded prior to analysis.
For each comparison and question, we report the percentage of judgments preferring SGS (ties counted as half a preference) with a 95\% Wilson score confidence interval and a two-sided exact binomial test against the 50\% chance level (ties excluded from the significance test), alongside Krippendorff's $\alpha$ (nominal) as a measure of inter-rater reliability\cite{krippendorff2004measuring}; results are reported in Table~\ref{tab:human_eval}.

\end{document}